\documentclass[journal]{IEEEtran}
\usepackage{amsmath,graphicx}
\usepackage{color,hyperref}
\usepackage[table,xcdraw]{xcolor}

\usepackage{subfigure}
\usepackage{times}
\usepackage{epsfig}
\usepackage{graphicx}
\usepackage{amsmath}
\usepackage{amssymb}

\usepackage{algorithm}
\usepackage{algorithmicx}
\usepackage{algpseudocode}
\usepackage{multirow}
\usepackage{booktabs} 
\usepackage{url}
\usepackage{bm}
\usepackage{verbatim}
\usepackage{breakurl}
\usepackage{bbding}
\usepackage{tabu}

\usepackage{stfloats}
\definecolor{bad}{RGB}{68,196,4}

\newlength\savedwidth

\ifCLASSINFOpdf
\else
\fi
\begin{document}
%
\title{CM-Net: Concentric Mask based Arbitrary-Shaped Text Detection}
%
%
%

\author{Chuang~Yang,
		Mulin~Chen,
		Zhitong~Xiong,
		Yuan~Yuan,~\IEEEmembership{Senior Member,~IEEE},
		and~Qi~Wang,~\IEEEmembership{Senior Member,~IEEE}
\thanks{This work was supported by the National Natural Science Foundation of China under Grant U21B2041, U1864204, 61632018, 62106182, and 61825603.

Chuang~Yang is with the School of Computer Science, and with the School of Artificial Intelligence, Optics and Electronics (iOPEN), Northwestern Polytechnical University, Xi'an 710072, Shaanxi, P. R. China.

Mulin~Chen, Zhitong~Xiong, Yuan~Yuan, and Qi~Wang are with the School of Artificial Intelligence, Optics and Electronics (iOPEN), Northwestern Polytechnical University, Xi'an 710072, Shaanxi, P.R. China.

E-mail: cyang113@mail.nwpu.edu.cn, chenmulin@mail.nwpu.edu.cn, xiongzhitong@gmail.com, y.yuan.ieee@gmail.com, crabwq@gmail.com. }
\thanks{Qi~Wang is the corresponding author.}
}

%
%

\markboth{}%
{Shell \MakeLowercase{\textit{et al.}}: Bare Demo of IEEEtran.cls for IEEE Journals}
%



\maketitle

\begin{abstract}
	Recently fast arbitrary-shaped text detection has become an attractive research topic. However, most existing methods are non-real-time, which may fall short in intelligent systems. Although a few real-time text methods are proposed, the detection accuracy is far behind non-real-time methods. To improve the detection accuracy and speed simultaneously, we propose a novel fast and accurate text detection framework, namely CM-Net, which is constructed based on a new text representation method and a multi-perspective feature (MPF) module. The former can fit arbitrary-shaped text contours by concentric mask (CM) in an efficient and robust way. The latter encourages the network to learn more CM-related discriminative features from multiple perspectives and brings no extra computational cost. Benefiting the advantages of CM and MPF, the proposed CM-Net only needs to predict one CM of the text instance to rebuild the text contour and achieves the best balance between detection accuracy and speed compared with previous works. Moreover, to ensure that multi-perspective features are effectively learned, the multi-factor constraints loss is proposed. Extensive experiments demonstrate the proposed CM is efficient and robust to fit arbitrary-shaped text instances, and also validate the effectiveness of MPF and constraints loss for discriminative text features recognition. Furthermore, experimental results show that the proposed CM-Net is superior to existing state-of-the-art (SOTA) real-time text detection methods in both detection speed and accuracy on MSRA-TD500, CTW1500, Total-Text, and ICDAR2015 datasets.
\end{abstract}


%
\IEEEpeerreviewmaketitle

\section{Introduction}
\label{intro}
%
%
%
%
\IEEEPARstart{T}{ext} detection has become increasingly important and popular, which provides fundamental information for intelligent devices to understand scenes. Recent progresses made in object detection~\cite{lin2017feature,lin2017focal,yuan2019vssa}, semantic segmentation~\cite{milletari2016v,long2015fully,xiong2021ask}, and instance segmentation~\cite{he2017mask,xie2020polarmask,zhang2019mask} improve the development of text detection significantly. 

\begin{figure}
	\centering
	\subfigure{
		\begin{minipage}[b]{0.95\linewidth}
			\includegraphics[width=1\linewidth]{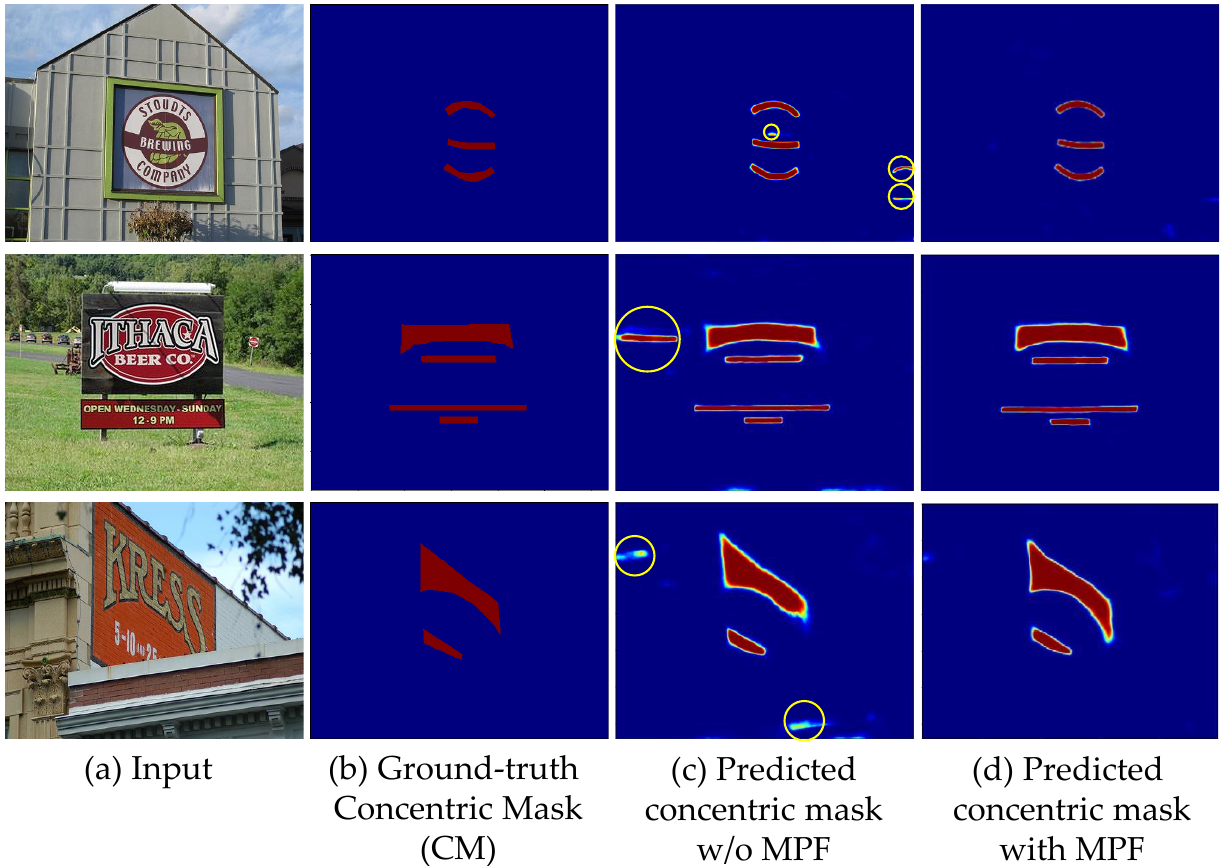}
	\end{minipage}}
	\vspace{-3mm}
	\caption{Comparison with ground-truth concentric masks and predicted concentric masks. (a) are the input RGB images. (b), (c), and (d) are the ground-truth concentric masks, predicted concentric masks without multi-perspective feature (MPF) module, and predicted concentric masks with MPF respectively. (c) fails to distinguish text regions from the background (the first and second rows yellow circles) and discriminate concentric masks from text regions (the third row yellow circles).}
	\label{V1}
	\vspace{-6mm}
\end{figure}

Recent research focus has shifted from multi-oriented text detection~\cite{zhou2017east,ma2018arbitrary,liao2018rotation} or arbitrary-shaped text detection~\cite{feng2019textdragon,wang2020textray,liu2020abcnet} to more challenging real-time arbitrary-shaped text detection~\cite{wang2019efficient,liao2020real}. Benefiting from the advantages of shrink masks~\cite{zhou2017east}, existing real-time text detection methods can represent text instances in a more efficient way, which brings significant improvements for detection speed. For example, Wang $et~al.$~\cite{wang2019efficient} modeled text instances by shrink masks, text masks, and similar vectors. They obtained the final text region by extending shrink masks to text masks based on similar vectors. Liao $et~al.$~\cite{liao2020real} extended shrink mask contours outward by fixed shrink distances to rebuild text contours. As for detection speed, these methods have fewer prediction headers and post-processing steps compared with previous works~\cite{zhang2020deep,ma2021relatext,wang2020contournet}, which speeds up the inference process. For detection accuracy, since they detect text instances based on shrink masks, the detection accuracy greatly relies on the recognition of shrink masks. 

However, the shrink mask is insensitive to text instance shapes, which leads to failure fitting for some irregular-shaped text contours and further influences the model performance. Moreover, three disadvantages exist in shrink masks based real-time arbitrary-shaped text detection methods, which heavily hamper them to recognize shrink masks accurately: (1) The deficient recognition of text contour features. Since the shrink mask is generated through moving text contours inward by a specific shrink distance, there is a strong correlation between text contours and the shrink mask contours. Therefore, the lack of recognition ability of text contours is adverse to recognizing shrink masks. (2) The patch-level text local features are neglected. It makes the model may suffer from the failure of discriminating text instances from the background. Specifically, these methods mainly learn the shrink mask features by training the networks by pixel-level classification tasks. It leads to small receptive fields for the model and further may result in false detection in some background regions that texture is similar to text instances. (3) The ambiguous definition of the gap region features between text contours and shrink mask contours, which hinders the recognition of shrink mask regions from text regions. Since both the gap regions and shrink masks are parts of text instances and they have similar texture features and semantic information, it is important for recognizing shrink masks to distinguish gap regions and shrink masks. But these methods do not explicitly define the difference between gap regions and shrink masks in the training stage, which makes the learned shrink mask features less discriminative.

Considering the aforementioned limitations, we propose a novel real-time text detector, namely CM-Net. Firstly, to fit arbitrary-shaped text contours more effectively, we propose a concentric mask (CM) to replace the shrink mask to represent text instances. Since the shrink distance of the CM considers text instance shapes, CM can fit text contours in an effective and robust manner. Secondly, as we mentioned before, the recognition of text border features, patch-level local features, and gap region features are helpful for the recognition of CM. Based on this finding, a multi-perspective feature (MPF) module is designed to extract these CM-related features, which brings significant improvement for recognizing CM (as shown in Fig.~\ref{V1}), and no extra computational cost to the inference process. At last, a multi-factor constraints loss is proposed to optimize the CM-Net. Benefiting from the aforementioned advantages, CM-Net further simplifies the framework and becomes the most effective and efficient text detection pipeline compared with existing state-of-the-art real-time arbitrary-shaped text detection methods (as shown in Fig.~\ref{V2}). The main contributions are as follows:

\begin{enumerate}
	\item A novel text representation model, Concentric mask (CM), is proposed to represent text instances in a more effective manner. Benefiting from its advantages, text detectors can more accurately fit some irregular-shaped text contours compared with the shrink mask~\cite{zhou2017east}.
	
	\item A multi-perspective feature (MPF) module is proposed to recognize CM accurately. It encourages the network to recognize CM from contour features, patch-level local features and gap features of text instances, and brings no extra computational cost to the inference process.
	
	\item A novel fast and accurate text detector is designed based on CM and MPF. In the inference stage, the network only needs to predict one CM to rebuild text contours. In the training process, MPF is helpful for the detector to predict CM more accurately. To the best of our knowledge, it is the most efficient framework among existing real-time arbitrary-shaped text detectors. 
	
	\item A multi-factor constraints loss is introduced to optimize the proposed detector for effectively recognizing CM. It balances multiple loss weights to focus on predicting CM. Moreover, it is insensitive to text scales, which improves the model robustness for various scales texts compared with existing loss such as the smooth-$l_1$ loss.
\end{enumerate}

The rest of the paper is organized as follows. Related works about text detection are shown in Section II. The details of our method are presented in Section III. The ablation experiments in Section IV to demonstrate the performance of the proposed method. In addition, we compared the CM-Net with its counterparts in Section V. Moreover, the detection results are visualized and detection speed is analyzed in Section VI. Finally, the conclusion of this paper is given in Section VII.

\section{Related Work}
\label{Related Work}
Existing text detection methods can be roughly divided into the regression-based methods, segmentation-based methods, and real-time arbitrary-shaped text detection methods. In this section, the related works are reviewed briefly.

\textbf{Regression-Based Text Detection Methods.} These approaches were inspired by objection detection frameworks. He $et~al.$~\cite{he2017single} designed a hierarchical inception module to extract the features with strong representation capacities and reduced the interference information from the background. To detect multi-oriented text instances accurately, Liao $et~al.$ \cite{liao2018textboxes++} designed an angle prediction header to predict the rotation angles between the horizontal anchors and multi-oriented text instances. Although the aforementioned works could effectively detect horizontal and multi-oriented text instances, they still suffered from the following limitations: (1) The settings of anchors depended on prior knowledge and the model performance was sensitive to it. (2) Plenty of anchors needed to be predefined to cover various texts, which made the model more complex and slowed down the detection speed largely. Thus, anchor-free text detection methods were proposed. Zhou $et~al.$ \cite{zhou2017east} predicted the offsets between the sample point and text border directly, which successfully avoided the influences that anchors brought to detectors. Liu $et~al.$ \cite{liu2019omnidirectional} remodeled texts with unordered points. It reduced the complexity of the features that the model needed to learn. 

To detect arbitrary-shaped text instances, Tang $et~al.$~\cite{tang2019seglink++} proposed a character-level text detection method. It detected multiple character boxes by SSD~\cite{liu2016ssd} at first. Then, it connected the characters belonging to the same text instances by attractive link and rejected the characters belonging to different text instances by repulsive link. Zhang $et~al.$~\cite{zhang2020deep} and Ma $et~al.$~\cite{ma2021relatext} also detected text instances in character-level. They predicted all potential character boxes at first. Then, the method generated linkage likelihoods of the boxes. In the end, they connected boxes by the corresponding linkages to obtain the text contours. Some other researchers~\cite{zhang2019look,wang2020contournet,wang2020all} generated proposal boxes to roughly locate text instances at first, and then sampling contour key points from the boxes. In the end, they reconstructed the text contours by connecting the sampling points in sequence. Wang $et~al.$~\cite{wang2020textray} obtained a contour points sequence by multiple rays directly and connected those points to get the final text contour, but it could not detect highly-curved texts effectively. Zhu $et~al.$~\cite{zhu2021fourier} represented text contours by Fourier Contour Embedding (FCE) method, which reconstructed text contours by predicting two score maps and Fourier signature vectors. Although the above approaches could detect irregular-shaped texts by improving traditional detection frameworks, the complicated framework heavily influenced model efficiency. 

\begin{figure}
	\centering
	\includegraphics[width=0.45\textwidth]{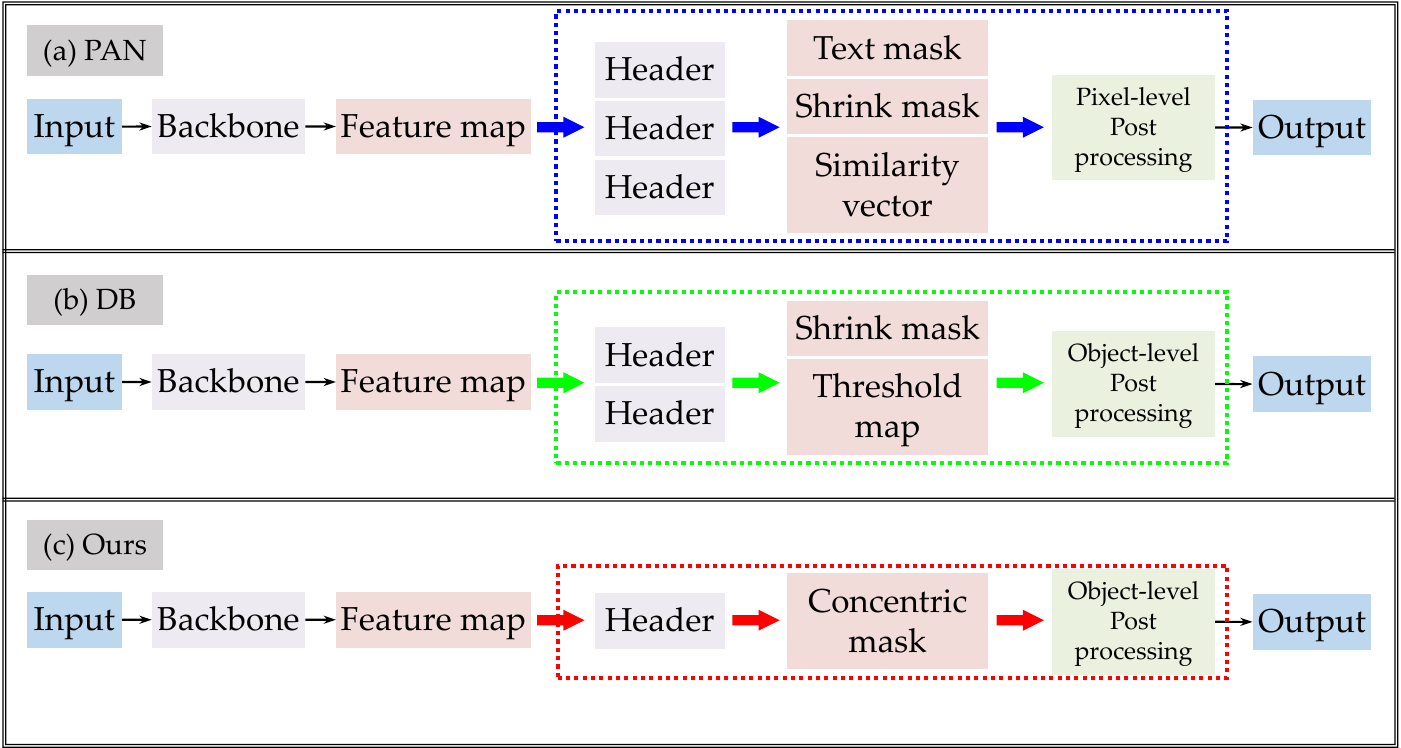}
	\vspace{-2mm}
	\caption{Comparison of whole pipelines of several representative real-time arbitrary-shaped text detectors. (a)-(c) are the frameworks of PAN, DB, and ours respectively.}
	\label{V2}
	\vspace{-6mm}
\end{figure}

\textbf{Segmentation-Based Text Detection Methods.} Since segmentation methods could fit any geometries tightly, many segmentation-based text detectors were proposed to detect arbitrary-shaped text instances. Lyu $et~al.$~\cite{lyu2018mask} roughly located the text instances through quadrilateral boxes at first. Then, the authors segmented text instance regions accurately from the boxes. Long $et~al.$~\cite{long2018textsnake} sampled a number of points on text center lines and predicted offsets between points and text contours. The text contours were rebuilt by applying offsets to points one by one. Baek $et~al.$~\cite{baek2019character} extended character-level datasets through weakly supervised networks to improve the model performance. Moreover, they proposed a 2D Gaussian character label to weak the text edge, which could separate the close characters. 

Following the idea of R-FCN~\cite{dai2016r}, Lyu $et~al.$~\cite{lyu2018multi} separated text instances into four different regions and combined them to get the final detection results. Zhang $et~al.$~\cite{zhang2020opmp} used an effective pyramid lengthwise and sidewise residual sequence modeling method to detect arbitrary-shaped text instances. Wang $et~al.$~\cite{wang2019shape} shrunk the text mask to generate a smaller mask, which could effectively separate adhesive texts and locate them roughly. In addition, the authors segmented the text mask and formed the final result by a pixel-level expansion process. Tian $et~al.$~\cite{tian2019learning} proposed a shape-aware embedding map for detecting text instances that accommodate various aspect ratios and imprecise boundaries, which is especially useful to the detection of long sentences. Jiang $et~al.$~\cite{jiang2020arbitrary} predicted text central region (TCR), expanding ratio, and text region (TR) at first. Then, the authors expanded TCR to TR by the expanding ratio and polygon expansion algorithm. Xu $et~al.$~\cite{xu2019textfield} proposed a text field to encode text instances. The magnitude and direction of the text field are used to locate text instances and separate adjacent text instances respectively. Although the aforementioned methods could detect arbitrary-shaped text instances effectively, they needed to predict plenty of text information by complicated network and handle them through many post-process steps, which heavily influenced text detection speed.

\textbf{Real-Time Arbitrary-Shaped Text Detection Methods.} Different from previous methods, real-time text detection methods focused on both detection accuracy and speed. Wang $et~al.$~\cite{wang2019efficient} roughly located text instances by shrink masks at first. Then, to rebuild text contours, they needed to extra predict text masks and similar vectors by the network to expand shrink mask regions to text regions with pixel-level processing (see Fig.~\ref{V2}~(a)). Liao $et~al.$~\cite{liao2020real} predicted shrink masks and the threshold maps simultaneously by the network (as shown in Fig.~\ref{V2}~(b)), the latter was used to refine the borders of shrink masks to improve the model detection accuracy. Compared with them, our method adopts a more efficient pipeline (as shown in Fig.~\ref{V2}~(c)). It only needs to predict one concentric mask by one prediction header, which makes our network more simple than PAN and DB. Moreover, CM-Net rebuilds text contours by object-level post-processing, which brings less computational cost than pixel-level post-processing of PAN.

\section{Our Method}
In this section, we first introduce the text representation method. Then, we illustrate the multi-perspective feature module in detail. Next, the overall framework of CM-Net is described. Moreover, the label generation process is visualized. In the end, the proposed multi-factor constraints loss and its salient properties are given.

\subsection{Text Representation Method}
\label{A}
The proposed method represents text instances by the concentric masks. Benefiting from the superiorities of concentric masks, arbitrary-shaped text contours can be fitted in a more robust way compared with existing state-of-the-art (SOTA) real-time text detection methods.

\begin{figure}
	\centering
	\includegraphics[width=0.45\textwidth]{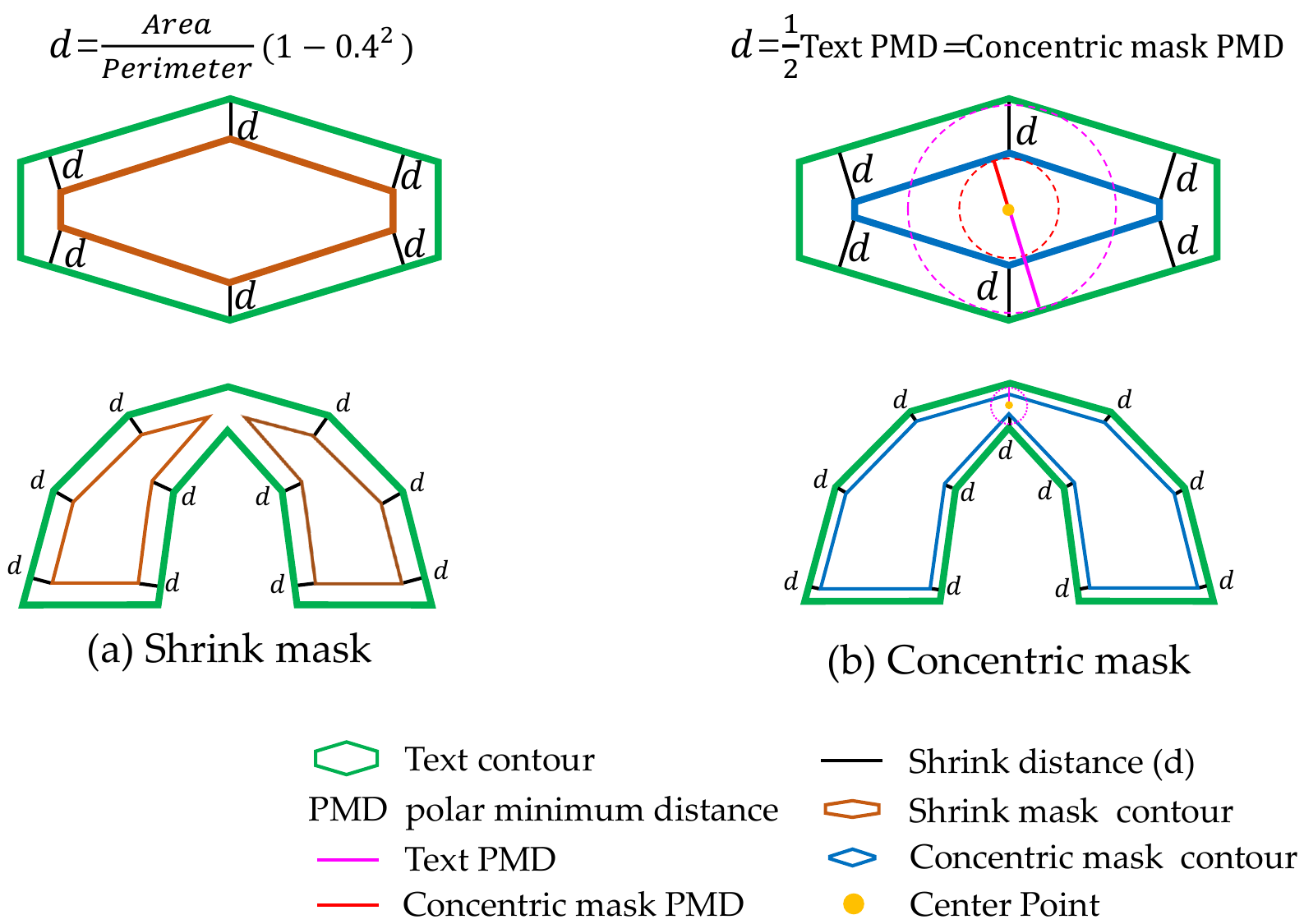}
	\vspace{-2mm}
	\caption{Visualization of shrink mask and concentric mask. (a) and (b) are the text representation results of shrink mask and concentric mask respectively. Shrink mask fails to represent some text instances (second row of (b)).}
	\label{V3}
\end{figure}

For the shrink mask~\cite{zhou2017east}, which is generated by moving text contours inward a specific shrink distance $d_{sm}$ that is computed by the following formula:
\begin{eqnarray} 
\label{dsm}
d_{sm}=\frac{S}{L}\left( 1-\sigma ^2 \right), 
\end{eqnarray} 
where $S$ and $L$ denote the area and perimeter of text instance. $\sigma$ is the shrink ratio. Since the $d_{sm}$ mainly relies on the global geometric characteristics of text instances, the $d_{sm}$-based shrink mask ignores the characteristic of local contour shape and fails to represent hourglass-shaped text instances. Specifically, as shown in Figure~\ref{V3}~second row~(a), the $d_{sm}$ is larger than the width of the narrow region, which leads to one text instance will correspond to multiple separated regions and breaks the integrity of the text instance.

\begin{figure}
	\centering
	\includegraphics[width=0.45\textwidth]{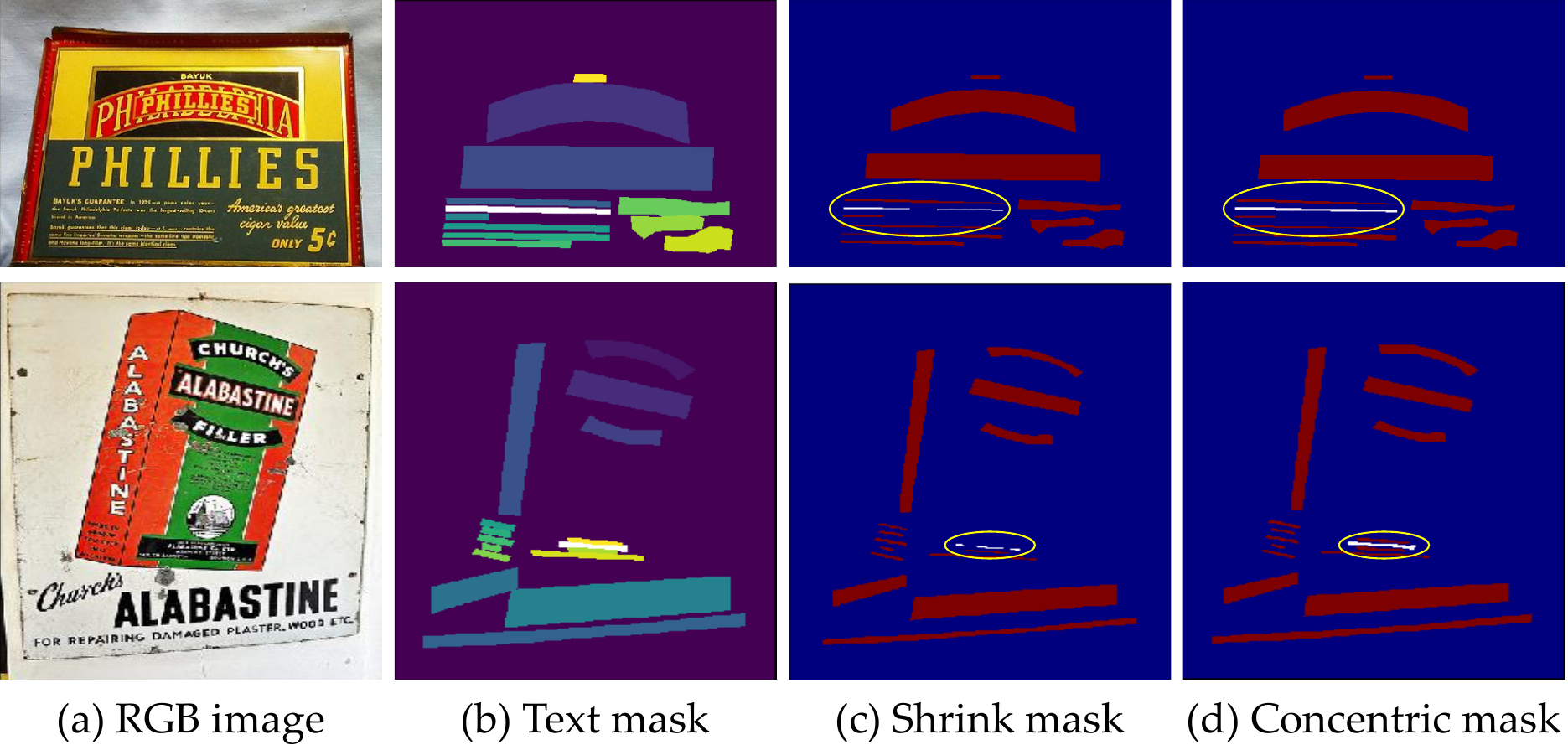}
	\vspace{-2mm}
	\caption{Demonstration of the limitation of shrink mask-based text representation method on CTW1500 dataset. (a) and (b) are the input RGB images and the corresponding ground-truth text masks. (c) and (d) are shrink masks and concentric masks that generated from (b). Shrink masks fail to represent some text instances (the white color regions in yellow circles).}
	\label{V3_1}
	\vspace{-4mm}
\end{figure}

\begin{figure}
	\centering
	\includegraphics[width=0.45\textwidth]{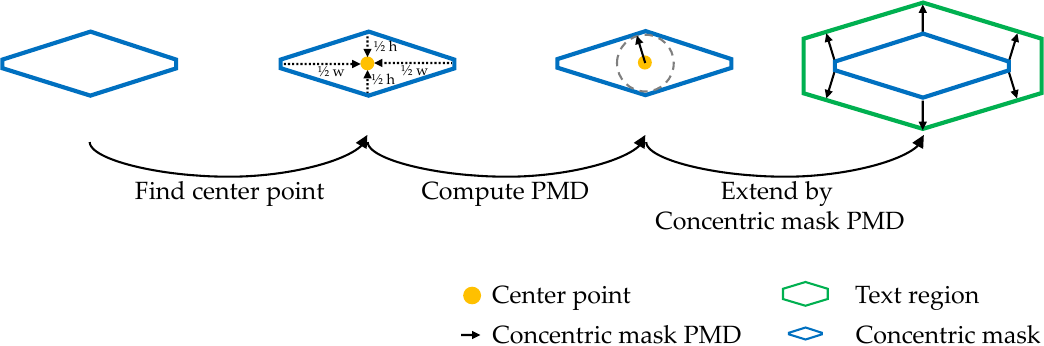}
	\vspace{-3mm}
	\caption{Visualization of reconstructing text contours by concentric masks.}
	\label{V4}
	\vspace{-6mm}
\end{figure}

Different from the shrink mask, the proposed concentric mask enjoys the same center point with text instance. At the same time, since the shrink distance $d_{cm}$ of concentric mask  considers the local geometric characteristic of text instance, the concentric mask effectively avoids representing one text instance by multiple separated regions (as shown in Fig.~\ref{V3}~second row of (b)), which enhances the model ability to fit irregular-shaped text instances in the real-world. The shrink distance $d_{cm}$ of the proposed concentric mask is defined as:
\begin{eqnarray} 
\begin{gathered}
d_{cm}=\frac{1}{2}PMD,\\
PMD={\rm min}(\left\| p_{cp},p_n \right\| _{2}^{2}),n=1,2,...,N,
\end{gathered}
\end{eqnarray} 
where $p_{cp}$ and $p_n$ are the center point and all contour points of text instance. $N$ is the number of contour points. $PMD$ is defined as the minimum distance between $p_{cp}$ and $p_n$. $\min \left( \cdot \right)$ and $||\cdot ||_2$ indicate the minimum operation and Euclidean distance, respectively. Compared with shrink mask, the $d_{cm}$-based concentric mask performs more robust for the representation of irregular-shaped text instances (as shown in Fig.~\ref{V3_1}). The process of reconstructing text contours by concentric masks is shown in Figrue~\ref{V4}.

\subsection{Multi-Perspective Feature Module}
\label{B}
As we mentioned before, CM-Net reconstructs text contours with concentric masks only. Therefore, the model detection accuracy heavily depends on the recognition of concentric masks. Since the recognition of concentric masks has a strong correlation with contour features, patch-level local features, and gap features of text instances, we propose a multi-perspective feature (MPF) module to encourage CM-Net to recognize concentric masks accurately. It consists of three prediction headers (polar minimum distance (PMD) prediction header, ray distance (RD) prediction header, and gap mask (GM) prediction header). They are helpful for learning concentric mask-related features and bring no extra computational cost to the inference process. 

\textbf{Polar minimum distance (PMD)} is defined as the shortest distance between the center point and contour in 360 degrees for any connected region. Therefore, after adding the PMD prediction header into the training stage, the network has to `look around' in 360 directions to recognize text local features before predicting PMD. It improves the model ability to recognize the patch-level text local features in the neighborhood of the center point, which brings significant improvement for the discrimination of text instances and the background regions that texture is similar to text instances.

\begin{figure*}
	\centering
	\includegraphics[width=0.85\textwidth]{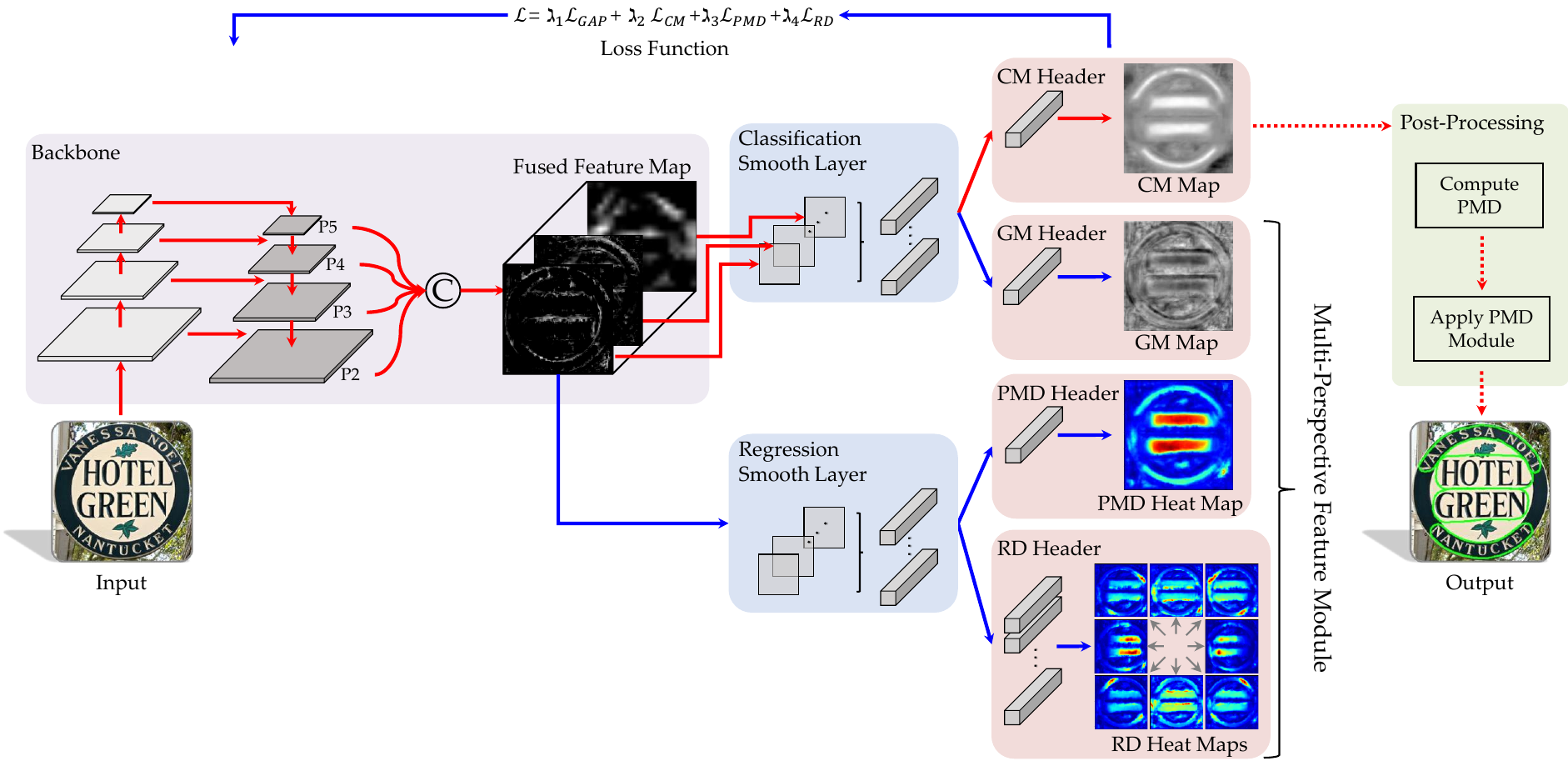}
	\vspace{-2mm}
	\caption{The overall framework of CM-Net. Blue solid flow and red solid flow are the training process, where blue flow indicates the multi-perspective feature module and it does not appear in the inference stage. Red dashed  arrows are the inference only operators. `C' means concatenation operation on multi-scale feature maps. Classification and regression smooth layers enjoy the same structure that includes a deep-wise CNNs layer with 3$\times$3 kernel and a normal CNNs layer with 1$\times$1 kernel. Post-processing is the text contours reconstructing process in Fig.~\ref{V4}.}
	\label{V5}
	\vspace{-4mm}
\end{figure*}

\textbf{Ray distance (RD)} denotes multiple distances between center point and text contour in different directions (such as in eight directions $0^{\circ}$, $45^{\circ}$, $90^{\circ}$, $135^{\circ}$, $180^{\circ}$, $225^{\circ}$, $270^{\circ}$, $325^{\circ}$). It means CM-Net has to `look forward' in multiple directions simultaneously to recognize text contours when training the network by RD prediction header, which improves the model ability to recognize the text contours and extends the network receptive fields simultaneously. Since concentric mask is generated by moving text contours inward by a specific distance, the recognition of text contour is critical for concentric mask detection.

\textbf{Gap mask (GM)} is the gap region between text contour and concentric mask contour, which is leveraged to train CM-Net to distinguish concentric mask regions from text regions. Since both concentric masks and gap masks are part of text instances, they enjoy similar high-level semantic context and low-level features (such as color, gradient, and texture), which makes it the model is hard to discriminate against them. To distinguish concentric mask regions from text regions accurately, gap mask regions are treated as a new semantic category different from text instances and the background in the training stage, which effectively avoids the ambiguity for recognizing concentric masks and gap masks from text regions and the background respectively.

The MPF module not only encourages CM-Net to recognize the concentric mask regions more accurately but also brings no extra computational cost to the inference process. It is the key component for CM-Net to detect arbitrary-shaped text instances with high detection accuracy and speed.
%
\subsection{Overall Framework}
In this paper, we propose a novel real-time text detector, it speeds up the inference process and improves the detection accuracy by multi-perspective feature module.

\begin{figure}
	\centering
	\includegraphics[width=0.45\textwidth]{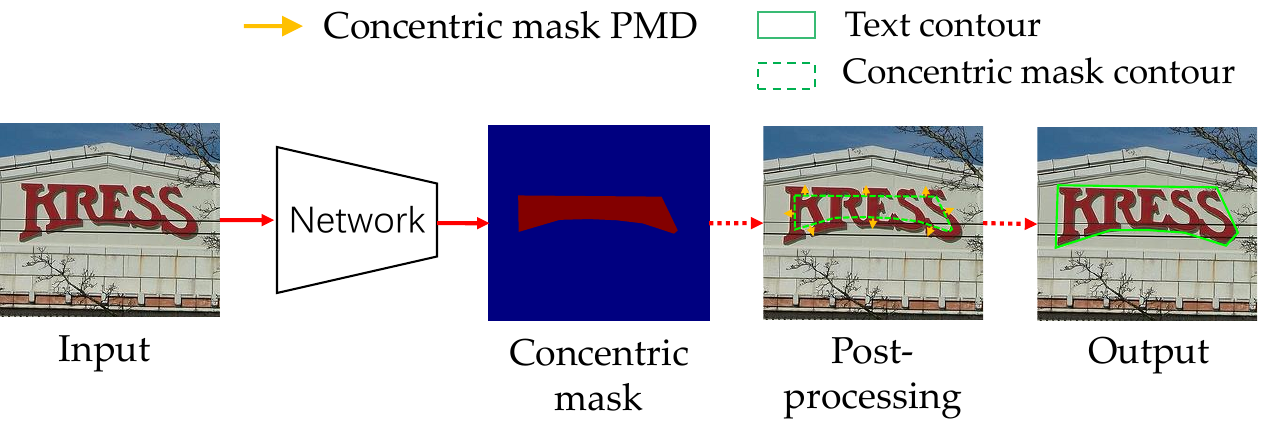}
	\vspace{-2mm}
	\caption{Inference pipeline of CM-Net. Network denotes the backbone, classification smooth layer, and CM header in Fig.~\ref{V5}. Post-processing is the text contours reconstructing process in Fig.~\ref{V4}.}
	\label{V6}
	\vspace{-4mm}
\end{figure}

\textbf{Network architecture} of the proposed CM-Net is shown in Fig.~\ref{V5}, which consists of backbone, smooth layers, concentric mask (CM) prediction header, multi-perspective feature module and post-processing. The backbone is combined by ResNet~\cite{he2016deep}, feature pyramid network (FPN)~\cite{lin2017feature} and concatenation layer ($C$). Multi-scale feature maps are extracted by ResNet and FPN at first. Then, they are concatenated by concatenation layer to generate a fused feature map. Smooth layers contain classification smooth layer and regression smooth layer. They enjoy the same network structure that consists of one 3$\times$3 group convolutional layer and one 1$\times$1 normal convolutional layer. Classification and regression smooth layers are utilized to provide classification smooth feature map and regression smooth feature map for the following CM prediction header and multi-perspective feature module respectively. CM prediction header conducts on the classification smooth feature map to segment CM for the post-processing. The multi-perspective feature module consists of gap mask (GM), polar minimum distance (PMD), and ray distance (RD) prediction headers, which are only used for training the proposed model to learn more discriminative CM-related features. Specifically, the GM prediction header conducts on the classification feature map to train classification smooth layer and backbone. PMD and RD prediction headers conduct on the regression feature map to train regression smooth layer and backbone. Particularly, CM, GM, and PMD prediction headers possess the same structure, which consists of one 3$\times$3 convolutional layer and one ReLU nonlinear activation layer. For RD prediction header, it is composed of eight 3$\times$3 convolutional layers and one ReLU nonlinear activation layer.

\begin{figure}
	\centering
	\includegraphics[width=0.45\textwidth]{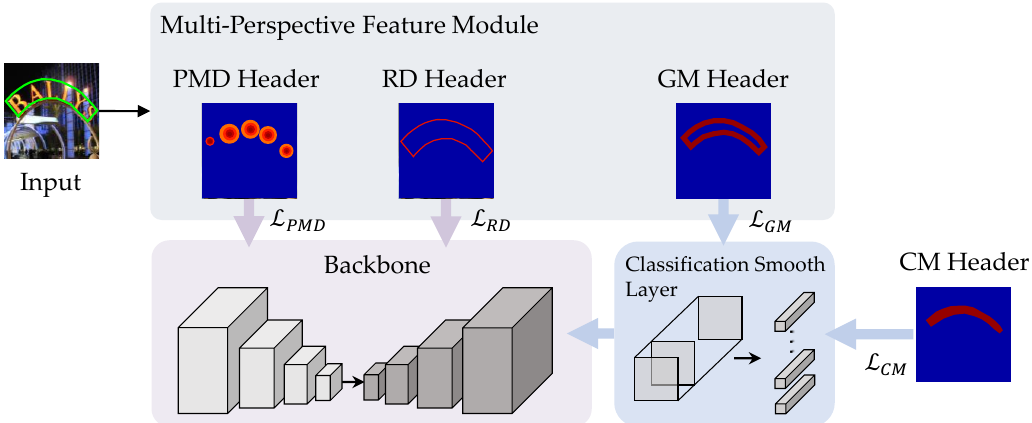}
	\vspace{-2mm}
	\caption{The details of training process. PMD and RD headers encourage the backbone to learn the text patch-level local features and text contour features respectively. The GM and CM headers optimize classification smooth layer and backbone to recognize the gap masks and concentric masks. ${\cal L}_{\rm RD}, {\cal L}_{\rm PMD}, {\cal L}_{\rm GM},$ and ${\cal L}_{\rm CM}$ are the objective functions of the RD, PMD, GM and CM headers respectively.}
	\label{V7}
	\vspace{-6mm}
\end{figure}

\begin{figure*}
	\centering
	\includegraphics[width=0.85\textwidth]{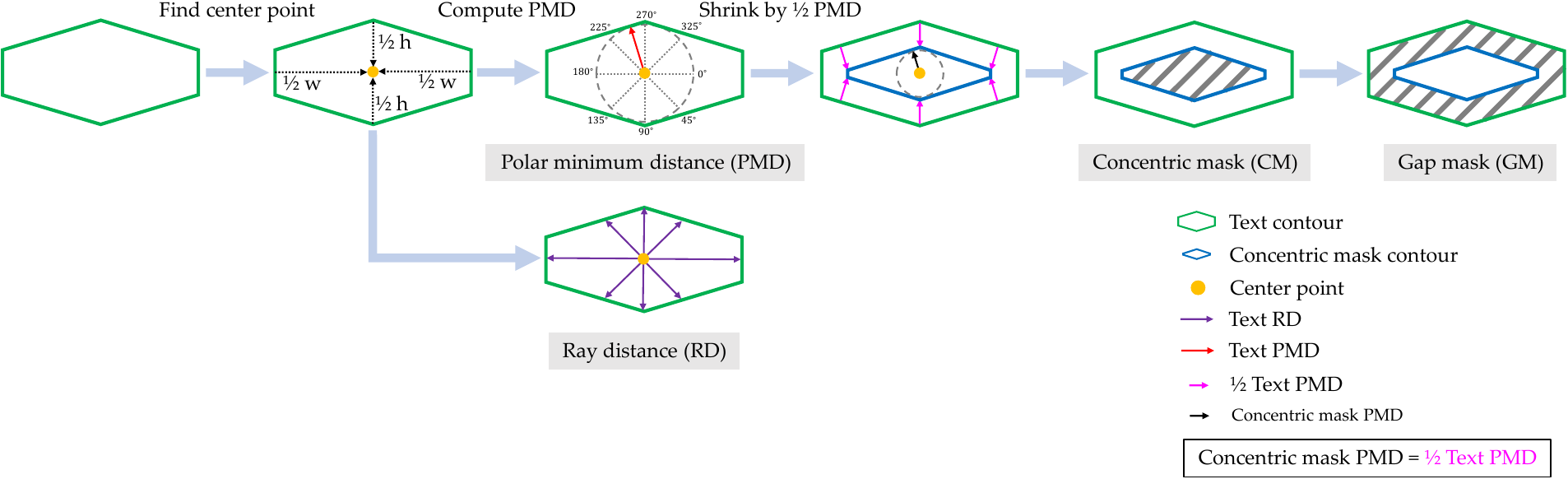}
	\vspace{-2mm}
	\caption{The illustration of label generation. PMD and RD are the distances computed by the text contour and center point. CM is the concentric mask region and GM is the gap region between text contour and concentric contour.}
	\label{V8}
	\vspace{-4mm}
\end{figure*}

\textbf{Inference flow} of the proposed method is illustrated in Fig.~\ref{V6}. Benefiting from the superiorities of the concentric mask (as mentioned in Section~\ref{A}), the CM-Net just needs to predict concentric masks when detecting text instances. Therefore, the network of CM-Net only needs to retain the flow of classification prediction header (red solid arrows in Fig.~\ref{V6}) in the inference process. The final text contours can be rebuilt by simple post-processing (as shown in Fig.~\ref{V4}), which mainly includes the three steps: (1) locating the center point; (2) computing the concentric mask PMD; (3) moving concentric mask contour outward by the concentric mask PMD.

\textbf{Training flow} of our method is shown in Fig.~\ref{V7}. Since multi-perspective feature module (solid blue arrows in Fig.~\ref{V5}) do not appear in the inference stage, the PMD and RD prediction headers mainly are used for optimizing the backbone of the network. The CM and GM prediction headers can optimize the backbone and classification smooth layer of the network simultaneously.

\subsection{Label Generation}
\label{lg}
The label generation processes of the concentric mask (CM), gap mask (GM), polar minimum distance (PMD), and ray distance (RD) are shown in Fig.~\ref{V8}. It mainly includes five steps: (1) Locating the center point of text contour. (2) Computing the text PMD and RD through text contour and the corresponding center point. (3) Moving text contour inward by $\frac{1}{2}$ text PMD to generate concentric mask contour. Note that, the concentric mask PMD equals $\frac{1}{2}$ text PMD, which is the key relationship between concentric mask and text. (4) Setting concentric mask region and other region to `1' and `0' respectively to generate concentric mask label. (5) Setting the gap mask region and other region to `1' and `0' to generate the gap mask label.

\begin{figure}
	\centering
	\includegraphics[width=0.45\textwidth]{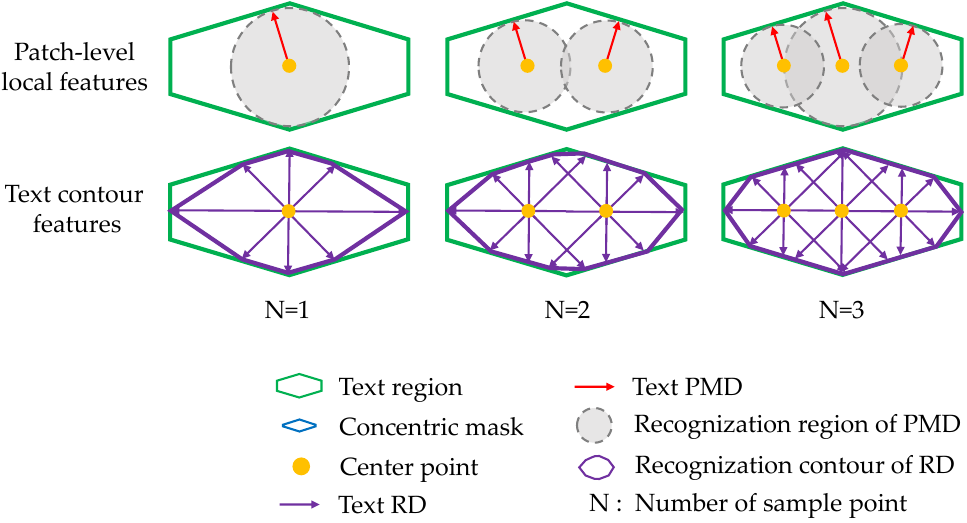}
	\vspace{-2mm}
	\caption{The visualization of PMD and RD with multiple sampling points. The first row and second row are the recognition area of PMD and RD respectively.}
	\label{V9}
	\vspace{-3mm}
\end{figure}

As we mentioned in Section~\ref{B}, training CM-Net by PMD and RD prediction headers can not only encourage the proposed method to recognize the text local features and text contour features but also extend the network receptive field. To fully exploit the advantages of PMD and RD features, we increase the number of sampling points in the label generation process. As shown in Fig.~\ref{V9}, with the increase of sampling points, the recognition area of PMD and the contour integrity of RD approximate the whole text region and contour, which can optimize the network better.

\subsection{Objective Optimization Function}
\label{oof}
In this paper, to optimize the the proposed pipeline, we propose a multi-factor constraints loss. It consists of four loss functions of CM prediction header loss ${\cal L}_{CM}$, GM prediction header loss ${\cal L}_{GM}$, PMD prediction header loss${\cal L}_{PMD}$, and RD prediction header loss ${\cal L}_{RD}$, which are applied to optimize the corresponding headers in the training stage.

For the CM prediction header, it aims to maximize the IoU between the predicted CM (${\rm CM}_{pred}$) and ground-truth CM (${\rm CM}_{gt}$). In this paper, dice loss is employed to optimize this header, which can enhance the integrity between the optimization objective and the loss function compared with the cross-entropy loss function. The dice loss function is defined as follow:
\vspace{-2mm}
\begin{eqnarray}
{\cal L}_{dice}({{\rm M}_{pred}, {\rm M}_{gt}}) = 1-\frac{2\times| {\rm M}_{pred}\cap {\rm M}_{gt}|+1}{| {\rm M}_{pred} |+| {\rm M}_{gt}|+1},
\end{eqnarray}
where ${\rm M}_{pred}$ and ${\rm M}_{gt}$ are the predicted and ground-truth binary masks. The CM loss function based on the ${\cal L}_{dice}$ can be formulated by:
\vspace{-2mm}
\begin{eqnarray}
{\cal L}_{CM} = {\cal L}_{dice}({{\rm CM}_{pred}, {\rm CM}_{gt}}),
\end{eqnarray}
As we mentioned in Section~\ref{intro}, gap region features contribute highly to recognize CM from text regions. Therefore, to supervise the model to effectively extract text gap features by GM header, we need to maximize the IoU between the predicted gap mask (${\rm GM}_{pred}$) and the corresponding label ${\rm GM}_{gt}$. Because of the small samples of the text gap mask features, only the valid region ${\rm GM}_{valid}$ in ${\rm GM}_{pred}$ participate in the training process, and it can be computed by:
\begin{eqnarray}\label{eqvalid}
{\rm GM}_{valid}=\begin{cases}
{\rm GM}_{pred},&{\rm GM}_{gt}=1\\
0,&		{\rm GM}_{gt}=0\\
\end{cases},
\end{eqnarray}

After obtained the valid region ${\rm GM}_{valid}$, the GM loss ${\cal L}_{GM}$ is defined as follow:
\vspace{-2mm}
\begin{eqnarray}
{\cal L}_{GM} = {\cal L}_{dice}({{\rm GM}_{pred}^{valid}, {\rm GM}_{gt}^{valid}}),
\end{eqnarray}
where ${\cal L}_{GM}$ forces the network to learn the features of valid regions and ignore the `0' regions. The network is not influenced by the background and focuses on the gap mask regions.

Though CM and GM features are effective for representing the concentric masks, they may suffer from the ambiguity recognition problem. Thus, in this work, we choose to learn the text patch-level local features and text contour features by PMD and RD headers to obtain more robust representations for concentric mask detection. Since predicting PMD and RD are regression tasks, for improving the model robustness to different scales texts and speeding up the training process, the ratio loss function is employed in this paper. It treats the difference between the predicted and ground-truth offsets as a ratio, and the value range of it always is 0-1 no matter the text scale is small or big, which makes the model is easy to converge and keep the same sensitivity for all scales of texts. The ratio loss function can be formulated as:
\begin{eqnarray}
{\cal L}_{ratio}({{\rm y}_{pred}, {\rm y}_{gt}}) = \log \frac{\max( {\rm y}_{pred}, {\rm y}_{gt})}{\min( {\rm y}_{pred}, {\rm y}_{gt})},
\end{eqnarray}
where ${\rm y}_{pred}$ and ${\rm y}_{gt}$ are predicted offset and the corresponding ground-truth respectively. The PMD and RD headers loss functions based on ${\cal L}_{ratio}$ can be defined as:
\begin{eqnarray}
{\cal L}_{\rm PMD} = {\cal L}_{ratio}({{\rm PMD}_{pred}, {\rm PMD}_{gt}}),
\end{eqnarray}
\begin{eqnarray}
{\cal L}_{RD} = \frac{1}{m}\sum\nolimits_{k=1}^m{{\cal L}_{ratio}({{\rm RD}_{pred}^{k}, {\rm RD}_{gt}^{k}})},
\end{eqnarray}
where ${\cal L}_{RD}$ considers the offsets in $k$ directions simultaneously, which enhances the text edge recognition.

The overall loss function for training the proposed framework is a multi-task loss, which consists of four terms: 1) the CM loss ${\cal L}_{CM}$; 2) the GM loss ${\cal L}_{\text{GM}}$; 3) the PMD loss ${\cal L}_{PMD}$; 4) the RD loss ${\cal L}_{RD}$. It can be represented as
\begin{equation}
\begin{aligned}
{\cal L} = \lambda_1 {\cal L}_{CM} + \lambda_2 {\cal L}_{GM} + \lambda_3 {\cal L}_{\text{PMD}} + \lambda_4 {\cal L}_{RD},
\end{aligned}
\end{equation}
where $\lambda_1$, $\lambda_2$, $\lambda_3$ and $\lambda_4$ are balancing weights of loss components for the multi-perspective feature loss function.

\section{Experiments}
In this section, we introduce four representative public benchmarks and evaluation metrics at first. Then, the implementation details are illustrated. In the end, we analyze the model performance in detail.

\subsection{Datasets and Evaluation Metrics}
\textbf{Datasets.} Four popular scene text detection datasets are used to evaluate the proposed framework: (1) MSRA-TD500 \cite{yao2012detecting} dataset with multi-lingual, arbitrary-oriented, and long text lines. It contains 700 training images and 200 test images, where the training set includes 400 images from HUST-TR400~\cite{yao2014unified}. (2) CTW1500~\cite{yuliang2017detecting} has 1000 training images and 500 testing images. It is a recent challenging dataset for curve text detection and the text scales vary greatly. (3) Total-Text~\cite{ch2017total} is also a newly-released dataset for curve text detection. This dataset includes horizontal, multi-oriented, and curved text instances simultaneously, and consists of 1255 training images and 300 testing images. (4) ICDAR 2015~\cite{karatzas2015icdar}) dataset is proposed in ICDAR 2015 Robust Reading Competition, and it contains 1500 images, of which 1000 are used to train the model and the remaining  500 are taken as testing set. This dataset contains both horizontal and multi-oriented texts.

\textbf{Evaluation Metrics.} To make a fair comparison, we use P (precision), R (recall), F (f-measure), and FPS (Frames Per Second) for evaluation, where F can be computed by P and R. In the following experiments, F and FPS are used to evaluate the detection accuracy and speed of the model repspectively. Suppose FP, TP, and FN are the number of predicted False Positive, True Positive, and False Negative samples, and P, R, and F can be computed by the following equations:
\begin{eqnarray}
\begin{gathered}
P=TP/(TP+FP)\times 100\%, \\
R=TP/(TP+FN)\times 100\%, \\
F=(2 \times P \times R)/(P+R).
\end{gathered}
\end{eqnarray}

\subsection{Implementation Details}
ResNet~\cite{he2016deep} and Feature Pyramid Network (FPN)~\cite{lin2017feature} are applied as the model backbone. Data augmentation is used in our work including the following three strategies: (1) random horizontal flipping, (2) random scaling and cropping, (3) random rotating. Note that, CM-Net has not been pre-trained on any extra training datasets in all following experiments.

In the training stage, Adam \cite{kingma2014adam} is employed with an initial learning rate of 1e-6, and the learning rate is reduced by the ``poly'' learning rate strategy that proposed in~\cite{yu2018bisenet}, in which the initial rate is multiplied by $\left( 1-\frac{iter}{\max \_iter} \right) ^{0.9}$ in all experiments. We train CM-Net with batch size 16 on 2 GPUs for 500 epochs. As for the multi-task training, the parameters ${\lambda_1}$, ${\lambda_2}$, ${\lambda_3}$, and ${\lambda_4}$ are set to 1, 1, 0.25, and 0.25 for all experiments. All experimental results in this paper are tested by PyTorch with batch size of 1 on one 1080Ti GPU and one i7-6800k CPU in a single thread.

\begin{table*}[]
	\caption{The results of baseline with PMD and RD prediction headers in the different number of sampling points. ``PMD'' and ``RD'' denote polar minimum distance and ray distance respectively. ``baseline'' means the framework equipped with concentric mask prediction header only.}
	\vspace{-1mm}
	\centering
	\label{number}
	\begin{tabular}{c|c|ccc|ccc}
		\hline
		\#&           & PMD & RD & \begin{tabular}[c]{@{}c@{}}Number of \\ sampling points\end{tabular} & Precision & Recall & F-measure \\ \hline
		1 & baseline+ & \Checkmark   &    & 1                         & 85.4      & 76.3   & 80.6      \\ 
		2 & baseline+ & \Checkmark   &    & 3                         & 85.5      & 78.6   & 81.9      \\ 
		3 & baseline+ & \Checkmark   &    & 5                         & 83.8      & 81.2   & \textbf{82.6}       \\ 
		4 & baseline+ & \Checkmark   &    & 7                         & 84.5      & 80.6   & 82.5      \\ \hline
		5 & baseline+ &     & \Checkmark  & 1                         & 87.8      & 77.8   & 82.5      \\ 
		6 & baseline+ &     & \Checkmark  & 3                         & 87.5      & 78.4   & 82.7      \\ 
		7 & baseline+ &     & \Checkmark  & 5                         & 86.7      & 79.3   & \textbf{82.8}       \\ 
		8 & baseline+ &     & \Checkmark  & 7                         & 86.0      & 79.8   & 82.8      \\ \hline
	\end{tabular}
	\vspace{-3mm}
\end{table*}

\subsection{Model Analysis}
To analyze the performance of the proposed framework comprehensively, we do the following ablation studies with ResNet18 and FPN backbone on the MSRA-TD500 dataset. Note that, in these experiments, all models are trained without external datasets. The short sides of test images in MSRA-TD500 are set to 736.

\textbf{Influence of the Number of Sampling Points.} As shown in Fig.~\ref{V10}, with the increase of sampling points, the recognition area of polar minimum distance (PMD) and ray distance (RD) become bigger. To study the effect of the number of sampling points to PMD and RD, we vary it from 1 to 7 in Table~\ref{number}. Note that, `baseline' denotes the framework equipped with CM header only. 

Table~\ref{number}~\#1~--~\#4 are the experimental results of baseline with PMD header, we can find that the F-measure on the test set keeps rising with the growth of the number and begins to level off when the number $\geqslant$ 5. It mainly because the recognition area of PMD approximates the whole text region when the number $=$ 5. The same conclusion (see Table~\ref{number}~\#5~--~\#8) is also applicable to the baseline with RD header. Although the F-measures are equal when the number is set to 5 and 7, the latter needs a larger GPU memory in the training process. Therefore, we set the number of center points to 5 in all following experiments.

\textbf{Effectiveness of Multi-Perspective Feature Module.} We design multiple groups of experiments to verify the effectiveness of different prediction headers in multi-perspective feature (MPF) module.

As we can see from Table~\ref{mpf}, the MPF from the framework is removed to obtain the performance of baseline at first, which achieves 80.5 in F-measure (Table~\ref{mpf}~\#1). Then, we compare the baseline with `baseline~+~PMD', `baseline~+~RD', and `baseline~+~GM' in Table~\ref{mpf}~\#2~\#3~\#4. They can make about 2.1\%, 2.3\%, and 2.4\% improvements on F-measure respectively, which verifies the validity of each single prediction header in MPF. Next, we test the performance of `baseline~+~PMD~+~RD', it achieves 83.8\% in F-measure (see Table~\ref{mpf}~\#5) and surpasses the performance of `baseline~+~PMD' and `baseline~+~RD' to a large extent, which verifies the effectiveness of the combination of PMD and RD prediction headers. In the end, CM-Net (`baseline~+~PMD~+~RD~+~GM') brings 4.5\% improvements in F-measure compared with baseline (see Table~\ref{mpf}~\#6), which indicates the MPF
significantly improve the model ability to detect text instances. 

\begin{table}[]
	\caption{The results of models with different prediction headers of multi-perspective feature module. ``PMD'', ``RD'', and ``GM'' denote polar minimum distance, ray distance, and gap mask respectively. ``baseline'' means the framework equipped with concentric mask prediction header only.}
	\vspace{-1mm}
	\centering
	\setlength{\tabcolsep}{1.4mm}
	\label{mpf}
	\begin{tabular}{c|c|ccc|ccc}
		\hline
		\# &           & PMD & RD & GM & Precision& Recall & F-measure \\ \hline
		1  & baseline  &     &    &    & 87.5           & 74.6        & 80.5           \\ 
		2  & baseline+ & \Checkmark   &    &    & 83.8           & 81.2        & 82.6           \\ 
		3  & baseline+ &     & \Checkmark  &    & 86.7           & 79.3        & 82.8           \\ 
		4  & baseline+ &     &    & \Checkmark  & 88.4           & 78.0        & 82.9           \\ 
		5  & baseline+ & \Checkmark   & \Checkmark  &    & 87.0           & 80.8        & 83.8           \\ 
		6  & CM-Net & \Checkmark   & \Checkmark  & \Checkmark  & 89.9           & 80.6        & 85.0           \\ \hline
	\end{tabular}
	\vspace{-4mm}
\end{table}

\textbf{Ratio Loss vs. Smooth-$l_1$ Loss.}
Multi-factor constraints loss consists of classification loss and regression loss, where regression loss can be implemented by ratio loss ${\cal L}_{ratio}$ and smooth-$l_1$. To verify the superiorities of ${\cal L}_{ratio}$, we use ${\cal L}_{ratio}$ and smooth-$l_1$ loss to train the model respectively.

\begin{figure}
	\centering
	\subfigure[Decline curve of PMD~loss implemented by Smooth-$_{l1}$ loss]{
		\begin{minipage}[b]{0.43\linewidth}
			\includegraphics[width=1\linewidth]{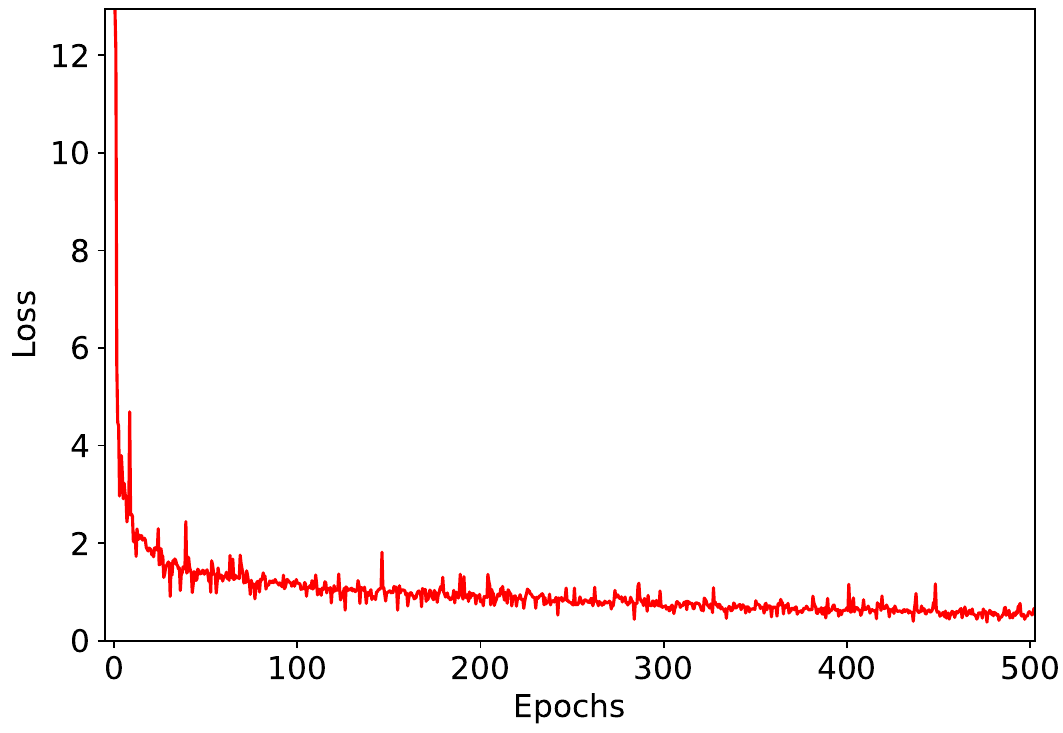}
	\end{minipage}}
	\vspace{-2mm}
	\subfigure[Decline curve of RD~loss implemented by Smooth-$_{l1}$ loss]{
		\begin{minipage}[b]{0.43\linewidth}
			\includegraphics[width=1\linewidth]{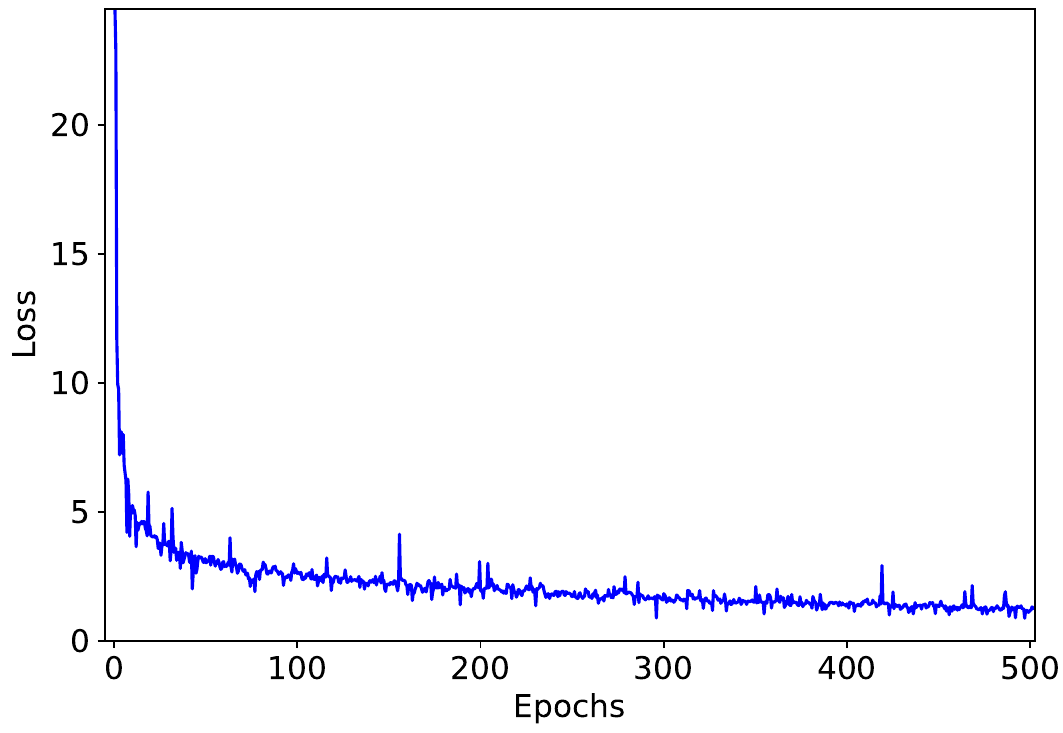}
	\end{minipage}}
	
	\subfigure[Decline curve of PMD~loss implemented by ratio loss]{
		\begin{minipage}[b]{0.43\linewidth}
			\includegraphics[width=1\linewidth]{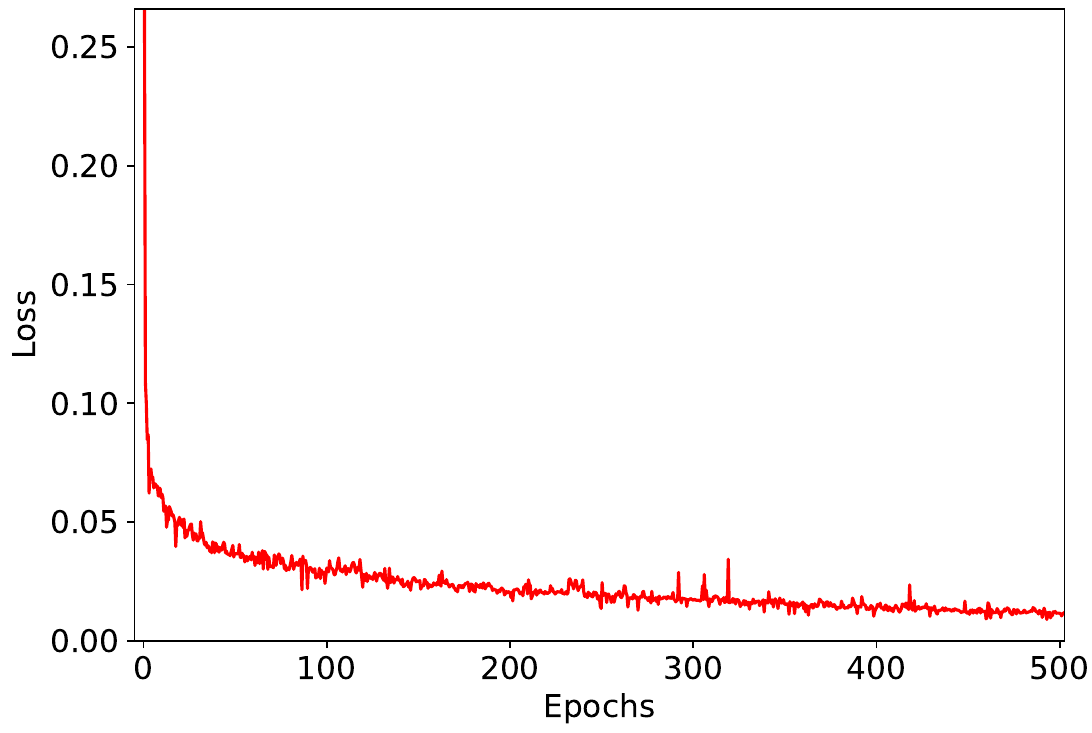}
	\end{minipage}}
	\subfigure[Decline curve of RD~loss implemented by ratio loss]{
		\begin{minipage}[b]{0.43\linewidth}
			\includegraphics[width=1\linewidth]{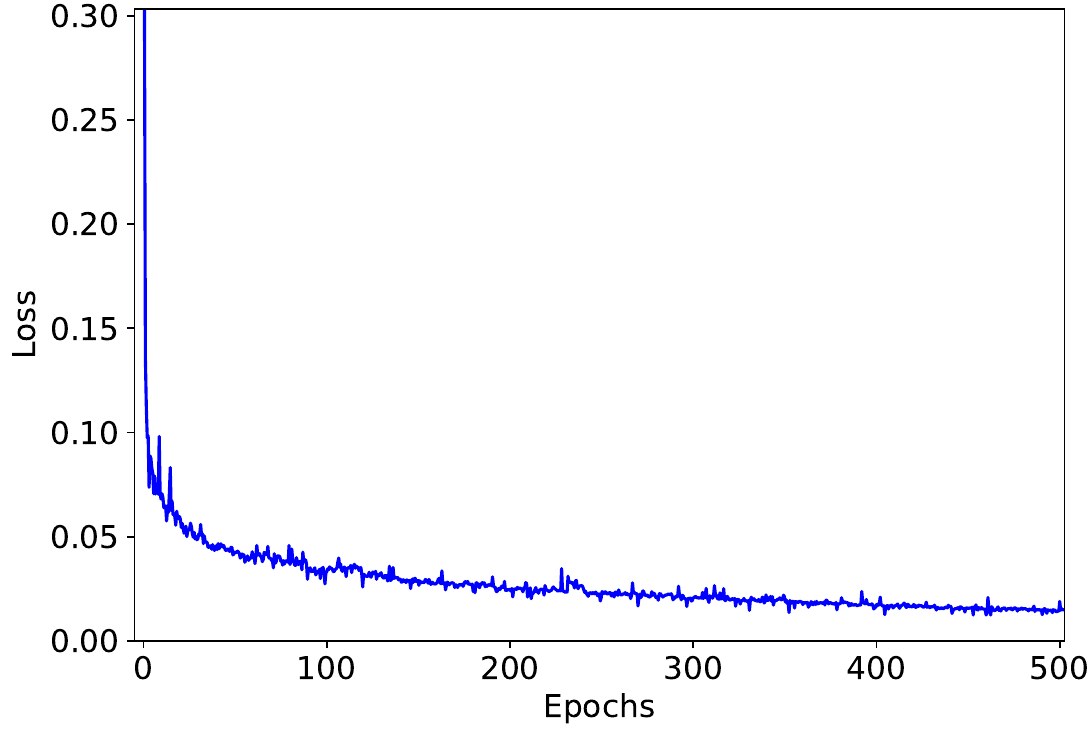}
	\end{minipage}}
	\vspace{-4mm}
	\caption{Decline curves of PMD~Loss and RD~Loss on MSRA-TD500.}
	\label{declinecurve}
	\vspace{-2mm}
\end{figure}

As illustrated in Fig.~\ref{declinecurve}, although bigger gradients from Smooth-$l_1$ speeds up the converging process (Fig.~\ref{declinecurve}~(a),(b)), they break out the training balance of the model and influence the detection accuracy deeply. As we can see from the Table~\ref{loss}, the ${\cal L}_{ratio}$ can bring 3.6\% improvement in F-measure compared with Smooth--$l_1$. It mainly because ${\cal L}_{ratio}$ limits loss weights to 0$\sim$1, which can bring the following three advantages compared to Smooth--$l_1$: (1) It makes the model is as insensitive to the text scale, which improves the CM-Net robustness for the detection of various scales texts; (2) It can balance the loss weights of multiple tasks and make the network focus on CM in the training stage. (3) It also smooths and stables the whole training process.

\textbf{Concentric Mask vs. Shrink Mask.} We also verify the effectiveness of the concentric mask. As shown in Table~\ref{represent}, the proposed CM-Net with the concentric mask can bring 0.8\% improvement in recall, which means the concentric mask is more effectively to fit arbitrary-shaped text instances compared with shrink mask (see Fig.~\ref{V3_1}). Moreover, the model with the concentric mask surpasses the model with the shrink mask about 0.3\% in F-measure, which demonstrates the concentric mask can bring better comprehensive performance.

\begin{table}[]
	\caption{The results of models trained by different loss functions. Regression loss denotes PMD loss and RD loss. ``Ratio'' is ratio loss in Section~\ref{oof}}
	\vspace{-1mm}
	\centering
	\setlength{\tabcolsep}{0.4mm}
	\label{loss}
	\begin{tabular}{c|cc|ccc}
		\hline
		\multirow{2}{*}{} & \multicolumn{2}{c|}{Regression loss} & \multirow{2}{*}{Precision } & \multirow{2}{*}{Recall } & \multirow{2}{*}{F-measure } \\ \cline{2-3}
		& Smooth-$l_1$           & Ratio          &                                 &                              &                                 \\ \hline
		CM-Net with               & \Checkmark                   &                & 90.7                            & 73.8                         & 81.4                            \\ 
		CM-Net with              &                     & \Checkmark              & 89.9                            & 80.6                         & 85.0                            \\ \hline
	\end{tabular}
	\vspace{-5mm}
\end{table}

\section{Comparison with State-of-the-Art Methods}
\label{cwem}
To show more details and give more comprehensive evaluations of the proposed work, we compare our work to other related existing state-of-the-art (SOTA) methods on MSRA-TD500, CTW1500, TotalText, and ICDAR2015 datasets. 

Existing SOTA methods can be divided into \textbf{Real-Time} and \textbf{Non-Real-Time} methods roughly. PAN~\cite{wang2019efficient} and DB~\cite{liao2020real} are SOTA real-time methods, they can detect arbitrary-shaped text instances with high detection accuracy and speed simultaneously. Since the proposed CM-Net is a real-time detector, we especially compared it with them to verify the superiority of our method in this section. Particularly, since DB with ResNet50 achieves better comprehensive performance, we compare our method with DB-ResNet50 in the following comparison experiments. Moreover, to ensure a fair comparison, we adopt the experimental results of SOTA methods not improved by Deformable Convolutional Network (DCN)~\cite{dai2017deformable}. 

\begin{table}[]
	\caption{The results of models with different text instance representation methods. ``SM'' and ``CM'' denote shrink mask and concentric mask respectively.}
	\centering
	\setlength{\tabcolsep}{0.6mm}
	\label{represent}
	\begin{tabular}{c|cc|ccc}
		\hline
		& SM& CM & Precision  & Recall  & F-measure  \\ \hline               
		CM-Net with               & \Checkmark                   &                & 90.2                            & 79.8                         & 84.7                            \\ 
		CM-Net with              &                     & \Checkmark              & 89.9                            & 80.6                         & 85.0                            \\ \hline
	\end{tabular}
	\vspace{-4mm}
\end{table}

\subsection{Evaluation on MSRA-TD500 Dataset}
To test the robustness of CM-Net to multiple languages, we evaluate the proposed method on MSRA-TD500 benchmark. To ensure a fair comparison, the short side of image is set to 736.

\textbf{Comparison with Real-Time Methods.} As shown in Table~\ref{msra}~MSRA-TD500, CM-Net is superior to all competitors in both detection accuracy and speed. Specifically, the proposed method achieves the F-measure of 85.0\% at an astonishing speed (41.7 FPS). For PAN~\cite{wang2019efficient} and DB~\cite{liao2020real}, our method outperforms them by 6.1\% and 3.1\% about F-measure, and 11.5 and 1.7 about FPS respectively.

Specifically, since the inference pipelines of PAN and DB have more headers compared with CM-Net (as shown in Fig.~\ref{V2}), it brings more computational cost (as illustration in Section~\ref{rvsa}) to the inference processes of PAN and DB and influences detection speed heavily. Moreover, benefiting from the proposed multi-perspective feature (MPF) module, CM-Net can learn more discriminative concentric mask-related features compared with PAN and DB, which encourages the proposed method to recognize text instances accurately and brings no extra computational cost to the inference process. Therefore, CM-Net can achieve better detection accuracy compared with PAN and DB while keeping high detection speed. For DB, although it improves the model detection accuracy by pre-training the network on SynthText~\cite{gupta2016synthetic}, CM-Net still brings 3.1\% improvements in F-measure (as we can see from Table~\ref{msra}~MSRA-TD500). This result demonstrates that the comprehensive performance of DB is not only far behind CM-Net but also heavily depends on plenty of training data, which limits the wide applications where the data is hard to sample.  

\begin{table}[]
	\caption{Comparison with related methods on MSRA-TD500, where ``Ext.'' denotes extra training data.}
	\label{msra}
	\centering
	\setlength{\tabcolsep}{0.8mm}
	\begin{tabular}{c|c|c|cccc}
		\hline
		\multicolumn{1}{c|}{Real-Time} & \multicolumn{1}{c|}{Methods} & \multicolumn{1}{c|}{Ext.} & \multicolumn{1}{c}{Precision} & \multicolumn{1}{c}{Recall} & \multicolumn{1}{c}{F-measure} & \multicolumn{1}{c}{FPS} \\ \hline
		&EAST~\cite{zhou2017east}                       & $\times$                         & 87.3                           & 67.4                        & 76.1                           & \textbf{13.2}                     \\
		&RRD~\cite{liao2018rotation}                 & \checkmark                         & 87.0                           & 73.0                        & 79.0                           & 10                     \\
		&PixelLink~\cite{deng2018pixellink}                 & \checkmark                         & 83.0                           & 73.2                        & 77.8                           & -                     \\
		&TextSnake~\cite{long2018textsnake}                 & \checkmark                         & 83.2                           & 73.9                        & 78.3                           & 1.1                      \\
		&CornerNet~\cite{lyu2018multi}                  & \checkmark                         & 87.6                           & 76.2                        & 81.5                           & 5.7                      \\
		No&CRAFT~\cite{baek2019character}                    & \checkmark                         & 88.2                           & 78.2                        & 82.9                           & 8.6                      \\
		&SAE~\cite{tian2019learning}                    & \checkmark                         & 84.2                          & 81.7                        & 82.9                           & -                      \\
		&TextField~\cite{xu2019textfield}                 & \checkmark                         & 87.4                           & 75.9                        & 81.3                           & -                        \\
		&SBD~\cite{liu2019omnidirectional}                        & $\times$                         & 89.6                           & 80.5                        & \textbf{84.8}                           & 3.2                      \\
		&ATTR~\cite{jiang2020arbitrary}                        & \checkmark                         & 84.2                           & 83.1                        & 83.6                           & -                      \\
		&OPMP~\cite{zhang2020opmp}                       & \checkmark                         & 86.0                           & 83.4                        & 84.7                           & 1.6                      \\ \hline
		&PAN~\cite{wang2019efficient})                        & $\times$                         & 80.7                           & 77.3                        & 78.9                           & 30.2                    \\
		Yes&DB~\cite{liao2020real}                         & \checkmark                         & 86.6                          & 77.7                        & 81.9                           & 40.0    
		\\ 
		&CM-Net (Ours)                   & $\times$                         & 89.9                           & 80.6                     &\textbf{85.0}                           & \textbf{41.7}      \\ \hline              
	\end{tabular}
	\vspace{-5mm}
\end{table}

\textbf{Comparison with Non-Real-Time Methods}. 
For some non-real-time methods such as SBD~\cite{liu2019omnidirectional}, ATTR~\cite{jiang2020arbitrary}, and OPMP~\cite{zhang2020opmp}, CM-Net not only outperforms them in F-measure but also runs 10x times faster than them. It mainly because these non-real-time methods model text instances from bottom-up perspectives, seeking to rebuild complex geometric with a series of local units and complicated post-processing, which makes the methods are hard to optimize and brings much computational cost to the inference process.

The performance on MSRA-TD500 demonstrates the solid superiority and robustness of the proposed CM-Net to detect multi-language long straight text instances.  Some qualitative illustrations on MSRA-TD500 are shown in Fig.~\ref{V10}~(f).
\begin{table*}[]
	\caption{Comparison with related methods on CTW1500 and Total-Text, where ``Ext.'' denotes extra training data. ``*'' means the results implemented by us.}
	\label{ctw1500tt}
	\centering
	\setlength{\tabcolsep}{1.2mm}
	\begin{tabular}{c|c|c|cccc|cccc}
		\hline
		\multicolumn{1}{c|}{\multirow{2}{*}{Real-Time}} &\multicolumn{1}{c|}{\multirow{2}{*}{Methods}} & \multicolumn{1}{c|}{\multirow{2}{*}{Ext.}} &  \multicolumn{4}{c|}{CTW1500}  & \multicolumn{4}{c}{Total-Text}                                                                                          \\ \cline{4-11} 
		\multicolumn{1}{c|}{}                         & \multicolumn{1}{c|}{}                      & \multicolumn{1}{c|}{}                      &\multicolumn{1}{c}{Precision } & \multicolumn{1}{c}{Recall } & \multicolumn{1}{c}{F-measure } & \multicolumn{1}{c|}{FPS} & \multicolumn{1}{c}{Precision } & \multicolumn{1}{c}{Recall } & \multicolumn{1}{c}{F-measure } & \multicolumn{1}{c}{FPS} \\ \hline
		&EAST~\cite{zhou2017east}                                        & $\times$                                          & 78.7                           & 49.1                        & 60.4                           & \textbf{21.2}                     & 50.0                           & 36.2                        & 42.0                           & -                        \\
		&TextSnake~\cite{long2018textsnake}                                   & \checkmark                                          & 67.9                           & 85.3                        & 75.6                           & 1.1                      & 82.7                           & 74.5                        & 78.4                           & -                        \\
		&CRAFT~\cite{baek2019character}                                       & \checkmark                                          & 86.0                           & 81.1                        & 83.5                           & -                        & 87.6                           & 79.9                        & 83.6                           & -                        \\
		&PSE~\cite{wang2019shape}                                            & $\times$                                           & 80.6                           & 75.6                        & 78.0                           & 3.9                      & 81.8                           & 75.1                        & 78.3                           & 3.9                      \\
		&LOMO~\cite{zhang2019look}                                           & \checkmark                                            & 85.7                           & 76.5                        & 80.8                           & -                        & 87.6                           & 79.3                        & 83.3                           & -                        \\
		&ATTR~\cite{jiang2020arbitrary}                                  & \checkmark                                          & 84.9                           & 80.3                        & 82.5                           & -                        & 89.0                           & 80.9                        & 84.8                           & -                        \\
		No&TextField~\cite{xu2019textfield}                                  & \checkmark                                          & 83.0                           & 79.8                        & 81.4                           & -                        & 81.2                           & 79.9                        & 80.6                           & -                        \\
		&ContourNet~\cite{wang2020contournet}                                        & $\times$                                           & 83.7                           & 84.1                        & 83.9                           & 4.5                        & 86.9                           & 83.9                        & \textbf{85.4}                           & 3.8                        \\
		&TextRay~\cite{wang2020textray}                                        & $\times$                                           & 82.8                           & 80.4                        & 81.6                           & -                        & 83.5                           & 77.9                        & 80.6                           & -                        \\
		&ABCNet~\cite{liu2020abcnet}                                        & \checkmark                                           & 81.4                           & 78.5                        & 81.6                           & -                        & 87.9                          & 81.3                        & 84.5                           & -                        \\
		&OPMP~\cite{zhang2020opmp}                                           & \checkmark                                           & 85.1                           & 80.8                        & 82.9                           & 1.4                      & 85.2                           & 80.3                        & 82.7                           & 3.7                      \\ 
		&FCENet~\cite{zhu2021fourier}                                           & $\times$                                           & 85.7                           & 80.7                        & 83.1                           & -                      & 87.4                           & 79.8                        & 83.4                           & -                      \\
		&ReLaText~\cite{ma2021relatext}                                           & \checkmark                                           & 84.8                           & 83.1                        & \textbf{84.0}                           & -                      & 86.2                           & 83.3                        & 84.8                           & \textbf{10.6}                      \\ \hline
		&PAN~\cite{wang2019efficient}                                        & $\times$                                          & 84.6                           & 77.7                        & 81.0                           & 39.8                     & 88.0                           & 79.4                        & 83.5                           & 39.6                     \\
		Yes&DB~\cite{liao2020real}                                          & \checkmark                                          & 84.3                           & 79.1                       & 81.6                           & 27.0                      & 86.9*                            & 79.7*                          & 83.1*                             & 27.5*                         \\
		&CM-Net (Ours)                                    & $\times$                                          & 86.0                           & 82.2                        & \textbf{84.1}                            & \textbf{50.3}                      & 88.5                           & 81.4                        & \textbf{84.8}                            & \textbf{49.8}         \\ \hline             
	\end{tabular}
\vspace{-4mm}
\end{table*}

\subsection{Evaluation on CTW1500 and Total-Text Datasets}
To evaluate the performance of our method for detecting curved and adhesive text instances, the proposed CM-Net is compared with other SOTA methods on CTW1500 and Total-Text datasets respectively. The short edges of test images in the two datasets are resized to 640.

\textbf{Comparison with Real-Time Methods}.
As we can see from Table~\ref{ctw1500tt}~CTW1500 and Total-Text, the conclusions on these two curved text benchmarks are the same as the conclusions on MSRA-TD500. On CTW1500 and Total-Text, CM-Net achieves 84.1\% and 84.8\% in F-measure, and 50.3 and 49.8 FPS in detection speed, which surpasses the performance of PAN~\cite{wang2019efficient} and DB~\cite{liao2020real} simultaneously to a large extent. Particularly, on CTW1500, CM-Net runs 10.5 FPS and 23.3 FPS faster than PAN and DB and outperforms them 3.1\% and 2.5\% in F-measure. On the Total-Text dataset, we get the detection result of the DB~\cite{liao2020real} without DCN through the official code to ensure the fairness of comparison. Similar conclusions can be concluded that the proposed CM-Net also performs better than PAN and DB both in detection accuracy and speed, which verifies the superiority of CM-Net.

\textbf{Comparison with Non-Real-Time Methods}.
On the CTW1500 dataset, CM-Net outperforms all SOTA methods both in detection accuracy and speed. A visualization example of the learned concentric masks and some involved post-processing steps are depicted in Fig.~\ref{V10}~(a)--(e).  The proposed CM-Net successfully detects concentric masks of arbitrary-shaped text instances (Fig.~\ref{V10}~(b)) and rebuilds text contour with simple post-processing (Fig.~\ref{V10}~(c),~(d))). Some qualitative illustrations are shown in Fig.~\ref{V10}~(g), which demonstrate our method can effectively detect extreme irregular-shaped geometries (highly-curved text instances) and adhesive text instances.

On the Total-Text dataset, although our method is a little lower (0.6\%) than ContourNet~\cite{wang2020contournet} in F-measure, CM-Net has a least 13 times faster speed (49.8 FPS) than it. In detection accuracy, ContourNet is following the architecture of Mask-RCNN~\cite{he2017mask}, it first roughly locates the text instances by quadrilateral boxes and then accurately segments the regions of text instances from the boxes, which avoids the interference brought by the background but complicated the framework and deeply influences the detection speed. Compared with ContourNet, since our method detect text instances from the background directly, CM-Net has great superiority in detection speed. Moreover, benefiting from the MPF module, the proposed CM-Net still can achieve competitive detection accuracy with ContourNet while bringing no influence to detection speed. At the same time, ReLaText~\cite{ma2021relatext} and ATTR~\cite{jiang2020arbitrary} need extra training data to train the network to achieve competitive detection accuracy with our method. They not only heavily depend on plenty of labeled data but also run slower than CM-Net 5 times at least. The proposed CM-Net achieves the best comprehensive performance both in detection accuracy and speed. Some qualitative results are shown in Fig.~\ref{V10}~(h).

The superior performance on CTW1500 and Total-Text demonstrates the CM-Net can successfully detect adhesive arbitrary-shaped text instances.

\subsection{Evaluation on ICDAR2015 Dataset}
We also verify the CM-Net ability to detect multi-oriented, small, and low-resolution text instances on the ICDAR2015 benchmark. During testing, we rescale the short side of input images to 736 like other SOTA methods.

\textbf{Comparison with Real-Time Methods}.
The model detection accuracy and speed on ICDAR2015 are shown in Table~\ref{ic15}. CM-Net outperforms PAN~\cite{wang2019efficient} and DB~\cite{liao2020real} by a large margin, i.e., 3.6\% and 0.7\% improvements in F-measure, and runs 8.4 FPS and 11.9 FPS faster than PAN and DB respectively. The performance of the proposed CM-Net surpasses PAN and DB on MSRA-TD500, CTW1500, Total-Text, and ICDAR2015 datasets simultaneously, which demonstrates our method is the best real-time arbitrary-shaped text detection method.

\begin{table}[]
	\caption{Comparison with related methods on ICDAR2015, where ``Ext.'' denotes extra training data. ``*'' means the results implemented by us.}
	\label{ic15}
	\centering
	\setlength{\tabcolsep}{0.8mm}
	\begin{tabular}{c|c|c|cccc}
		\hline
		\multicolumn{1}{c|}{Real-Time} &\multicolumn{1}{c|}{Methods} & \multicolumn{1}{c|}{Ext.} & \multicolumn{1}{c}{Precision} & \multicolumn{1}{c}{Recall} & \multicolumn{1}{c}{F-measure} & \multicolumn{1}{c}{FPS} \\ \hline
		&EAST~\cite{zhou2017east}                       & $\times$                         & 83.6                           & 73.5                        & 78.2                           & \textbf{13.2}                     \\
		&RRD~\cite{liao2018rotation}                 & \checkmark                         & 85.6                           & 79.0                        & 82.2                           & 6.5                     \\
		&TextSnake~\cite{long2018textsnake}                 & \checkmark                         & 84.9                           & 80.4                        & 82.6                           & 1.1                      \\
		&CornerNet~\cite{lyu2018multi}                 & \checkmark                         & 94.1                           & 70.7                        & 80.7                           & 3.6                      \\
		&PSE~\cite{wang2019shape}                  & $\times$                         & 81.5                           & 79.7                        & 80.6                           & 1.6                      \\
		&Boundary~\cite{wang2020all}                    & \checkmark                         & 88.1                           & 82.2                        & \textbf{85.0}                          & -                      \\
		No&TextField~\cite{xu2019textfield}                 & \checkmark                         & 84.3                           & 80.5                        & 82.4                           & 6.0                        \\
		&TextDragon~\cite{feng2019textdragon}                 & \checkmark                         & 84.8                           & 81.8                        & 83.1                           & -                        \\
		&SegLink++~\cite{tang2019seglink++}                 & \checkmark                         & 83.7                           & 80.3                        & 82.0                           & 7.1                        \\
		&ATTR~\cite{jiang2020arbitrary}                 & $\times$                         & 85.8                           & 79.7                       & 82.6                           & -                       \\
		&SAE~\cite{tian2019learning}                 & \checkmark                         & 85.1                           & 84.5                       & 84.8                           & 3.0                        \\
		&FCENet~\cite{zhu2021fourier}                       & $\times$                         & 85.1                          & 84.2                       & 84.6                           & -                      \\ \hline
		&PAN~\cite{wang2019efficient})                        & $\times$                         & 82.9                           & 77.8                        & 80.3                           & 26.1                    \\
		Yes&DB~\cite{liao2020real}                         & \checkmark                         & 86.5*                          & 80.2*                       & 83.2*                           & 22.6*     
		\\ 
		&CM-Net (Ours)                   & $\times$                         & 86.7                           & 81.3                     &\textbf{83.9}                           & \textbf{34.5}      \\ \hline              
	\end{tabular}
	\vspace{-4mm}
\end{table}

\begin{figure*}
	\centering
	\includegraphics[width=0.95\textwidth]{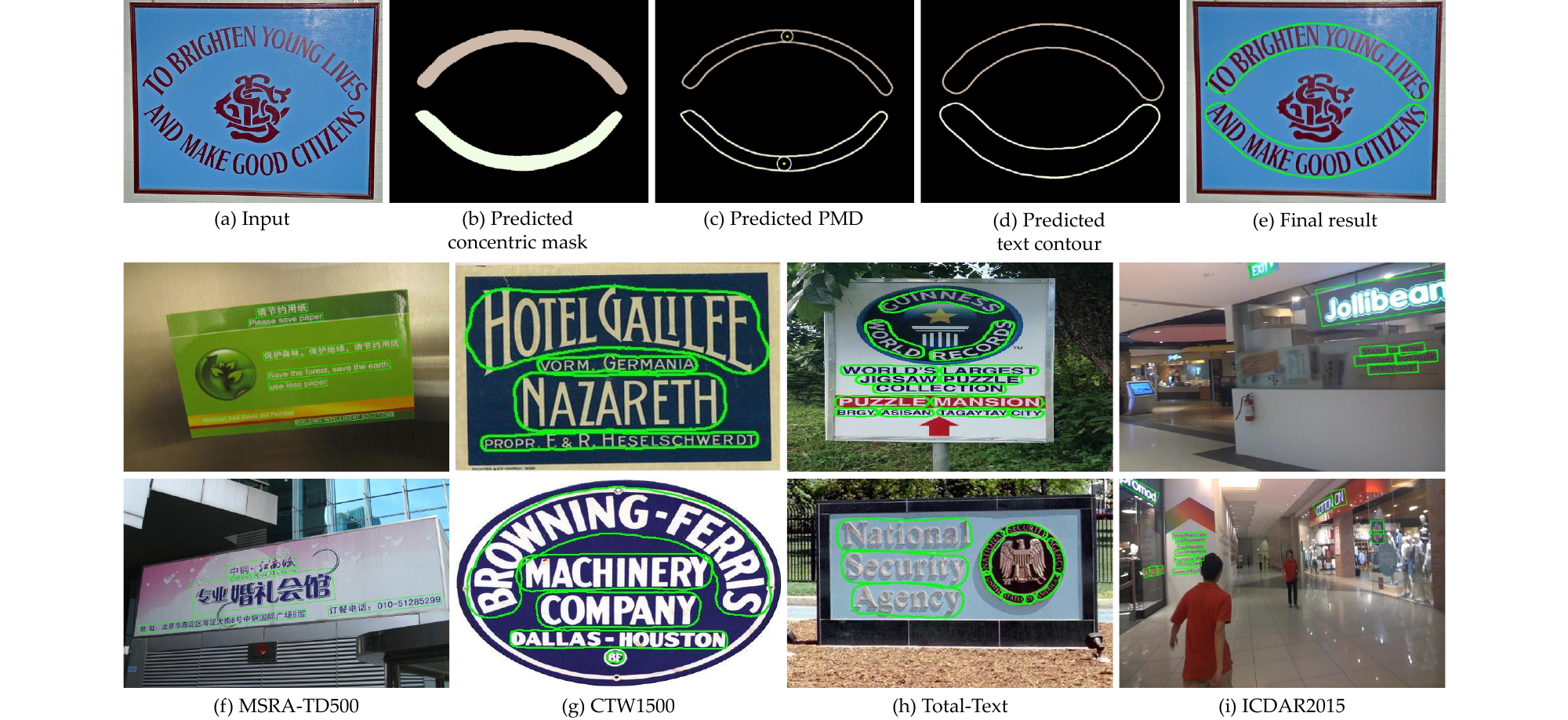}
	\vspace{-3mm}
	\caption{Qualitative detection results of CM-Net. (a) is the input image. (b) is the predicted concentric mask. (c) is the visualization of PMD, where the yellow solid circle is the center point and the radius of the small circle is the PMD. (d) is the reconstructed text contour by concentric mask and PMD. (e) is the final detection result of the input. (f)-(i) are detection results on four standard benchmarks.}
	\vspace{-4mm}
	\label{V10}
\end{figure*}

\textbf{Comparison with Non-Real-Time Methods}.
Compared with non-real-time methods, the proposed CM-Net not only enjoys absolute superiority in detection speed but also achieves competitive detection accuracy. As we can see from Table~\ref{ic15}, although the proposed CM-Net performs slightly worse than Boundary~\cite{wang2020all}, SAE~\cite{tian2019learning}, and FCENet~\cite{zhu2021fourier} in F-measure, our method runs 10x times faster than them. Specifically, for Boundary and SAE, the former refines the detection results by multiple detection stages, the latter extra predicts an embedding map to detect great aspect ratio text instances. Both of them complicated the frameworks and bring much computational cost to the inference process. At the same time, to achieve the best performance, they have to seek extra training data (SynthText~\cite{gupta2016synthetic}) to pre-train their network to optimize the huge network weights, which leads to heavy dependency on labeled data. For FCENet, it models text instances into the Fourier domain from the bottom to up perspective. In the inference process, it not only needs to predict plenty of text instance-related information but also have to transform them from the Fourier domain to the spatial domain and keep reliable results by non-maximum suppression (NMS)~\cite{rosenfeld1971edge}, which is more complicated than Boundary and SAE and far behind our method in detection speed.

The experiment results on ICDAR 2015 show that the proposed CM-Net can handle the texts with various scales and multi-orientations effectively. We show some qualitative detection results in Fig.~\ref{V10}~(i).

\begin{table}[]
	\caption{Cross-dataset evaluations on word-level and line-level datasets.}
	\vspace{-1mm}
	\label{cross}
	\centering
	\setlength{\tabcolsep}{4.3mm}
	\begin{tabular}{c|ccc}
		\hline
		& \multicolumn{3}{c}{Total-Text (train on ICDAR2015)}                                    \\ \cline{2-4} 
		\multirow{-2}{*}{Methods} & Precision              & Recall              & F-measure             \\ \hline
		CM-Net (Ours)                & 75.8 & 64.5 & 69.7 \\ \hline
		& \multicolumn{3}{c}{ICDAR2015 (train on Total-Text)}                                    \\ \cline{2-4} 
		\multirow{-2}{*}{Methods} & Precision               & Recall                 & F-measure             \\ \hline
		CM-Net (Ours)                & 76.5                        & 68.1                        & 72.1                        \\ \hline
		& \multicolumn{3}{c}{CTW1500 (train on MSRA-TD500)}                                      \\ \cline{2-4} 
		\multirow{-2}{*}{Methods} & Precision               & Recall                 & F-measure            \\ \hline
		CM-Net (Ours)                & 77.2 & 69.7 & 72.8 \\ \hline
		& \multicolumn{3}{c}{MSRA-TD500 (train on CTW1500)}                                      \\ \cline{2-4} 
		\multirow{-2}{*}{Methods} & Precision              & Recall              & F-measure             \\ \hline
		CM-Net (Ours)                & 85.8 & 77.1 & 81.2 \\ \hline
	\end{tabular}
\vspace{-5mm}
\end{table}

\vspace{-3mm}

\subsection{Cross Dataset Text Detection}
To demonstrate the generalization ability of the proposed CM-Net, we train our method on one dataset and test on a different dataset, where the training dataset and testing dataset are annotated at the same level (e.g., word-level or line-level).

Specifically, we design four groups of experiments on several classical datasets (MSRA-TD500, CTW1500, Total-Text, and ICDAR2015 datasets). Since the Total-Text and ICDAR2015 are word-level datasets and they consist of curved and multi-oriented text instances respectively, we first design two groups of cross experiments on them to verify the model generalization on word-level different shaped text instances. As we can see from Table~\ref{cross}, the proposed CM-Net achieves acceptable performance on the cross-dataset with Total-Text and ICDAR2015. Then, we also conduct two groups of cross experiments on CTW1500 and MSRA-TD500, which are line-level datasets and consist of curved and multi-oriented text instances respectively. As depicted in Table~\ref{cross}, CM-Net also achieves very competitive results. Particularly, on the MSTA-TD500, our method trained on CTW1500 performs better than some existing SOTA methods trained on MSTA-TD500 (e.g., EAST~\cite{zhou2017east}, RRD~\cite{liao2018rotation}, TextSnake~\cite{long2018textsnake}, and PAN~\cite{wang2019efficient}).

The experimental results not only demonstrate the proposed CM-Net enjoys strong generalization for different scene datasets and the robustness for different shaped text instances but also verify that CM-Net reduces the dependency on labeled data compared with other methods.

\begin{table}[]
	\centering
	\caption{Time consumption of CM-Net on four public benchmarks. The total time
		consists of backbone, segmentation head and post-processing. `Size' and `Post' represent the size of short edge of image and post-processing respectively.}
	\vspace{-1mm}
	\label{speed1}
	\setlength{\tabcolsep}{1.4mm}
	\begin{tabular}{c|c|ccc|cc}
		\hline
		\multirow{2}{*}{Dataset} & \multirow{2}{*}{Size} & \multicolumn{3}{c|}{Time consumption (ms)} & \multirow{2}{*}{F-measure} & \multirow{2}{*}{FPS} \\ \cline{3-5}
		&                       & Backbone        & Head        & Post       &                            &                      \\ \hline
		MSRA-TD500               & 736                   & 11.9            & 9.8         & 2.3        & 85.0                       & 41.7                 \\ 
		CTW1500                  & 640                   & 9.7             & 8           & 2.2        & 84.1                       & 50.3                 \\ 
		Total-Text               & 640                   & 10              & 7.8         & 2.3        & 84.8                       & 49.8                 \\ 
		ICDAR2015                & 736                   & 14.2            & 11.9        & 2.9        & 83.9                       & 34.5                 \\ \hline
	\end{tabular}
\vspace{-4mm}
\end{table}

\begin{table}[]
	\centering
	\caption{Comparison of computational cost and performance of different real-time detectors. `GFLOPs' indicate Floating Point of Operations.}
	\vspace{-1mm}
	\label{speed2}
	\setlength{\tabcolsep}{0.4mm}
	\begin{tabular}{c|c|cccc}
		\hline
		\multirow{2}{*}{Methods} & \multirow{2}{*}{GFLOPs} & \multicolumn{4}{c}{F-measure}    \\ \cline{3-6} 
		&                         & MSRA-TD500 & CTW1500 & Total-Text& ICDAR2015  \\ \hline
		PAN~\cite{wang2019efficient} & 63.88                   & 78.9       & 81.0    & 83.5  & 80.3      \\ 
		DB~\cite{liao2020real}                                      & 52.54                   & 81.9       & 81.6    & 83.1 & 83.2       \\ 
		CM-Net                          & 35.41                   & 85.0       & 84.1    & 84.8 & 83.9      \\ \hline
	\end{tabular}
\vspace{-4mm}
\end{table}

\section{Speed Analysis and Result Visualization}
\label{rvsa}
\textbf{Speed Analysis.}
\label{analyze}
We especially analyze the time consumption of CM-Net in different stages. As we can see from Table~\ref{speed1}, the backbone takes the most time, and the time cost of post-processing is about a quarter of the head. Additionally, the backbone and head will take more time with the increasing image size. At the same time, it is found that the time cost of post-processing is not sensitive to the scale of text instances (from Table~\ref{speed1}~first and second raws), which makes CM-Net perform better in scenes with a large text scale.

\begin{figure*}
	\centering
	\subfigure[vs. FCENet~\cite{zhu2021fourier}]{
		\begin{minipage}[b]{0.2\linewidth}
			\includegraphics[width=1\linewidth]{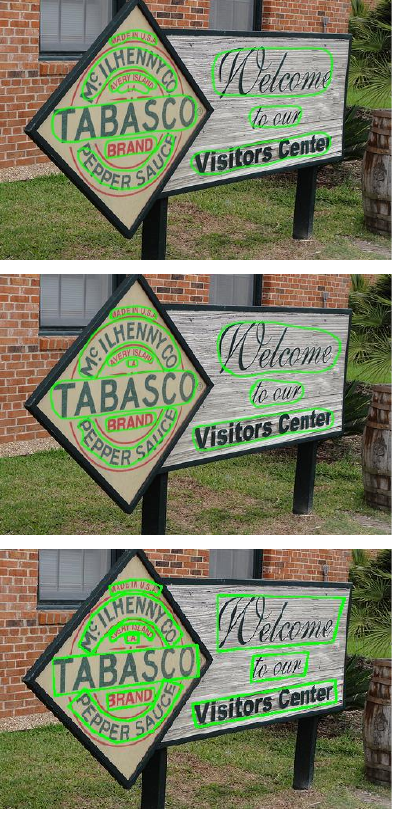}
	\end{minipage}}\hspace{-1.7mm}
	\subfigure[vs. TextField~\cite{xu2019textfield}]{
		\begin{minipage}[b]{0.2\linewidth}
			\includegraphics[width=1\linewidth]{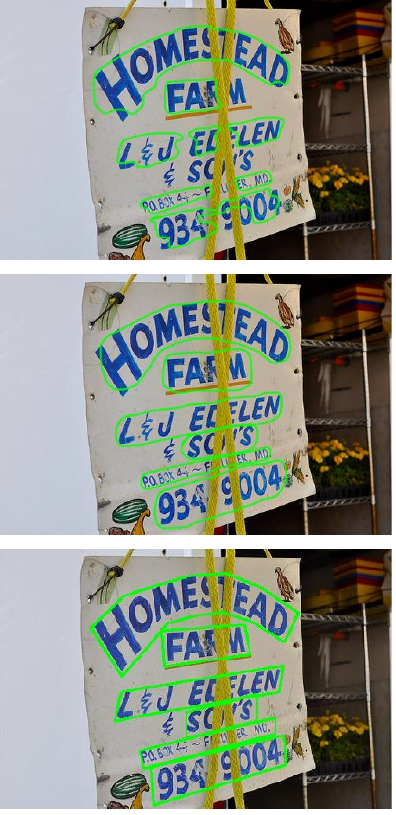}
	\end{minipage}}\hspace{-1.7mm}
	\subfigure[vs. OPMP~\cite{zhang2020opmp}]{
		\begin{minipage}[b]{0.2\linewidth}
			\includegraphics[width=1\linewidth]{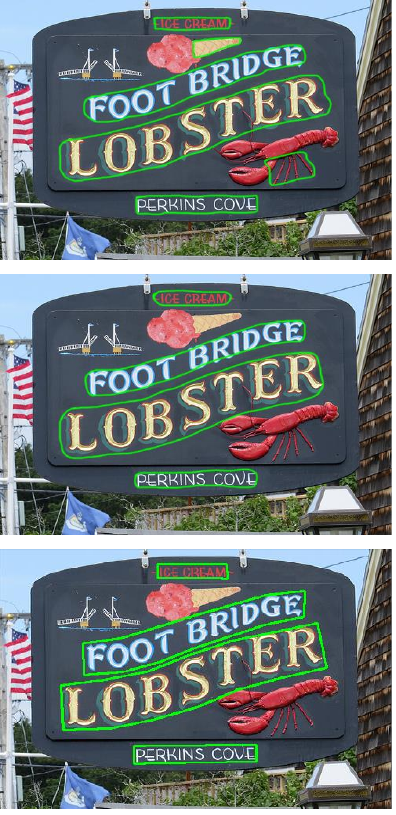}
	\end{minipage}}\hspace{-1.7mm}
	\subfigure[vs. ABCNet~\cite{liu2020abcnet}]{
		\begin{minipage}[b]{0.2\linewidth}
			\includegraphics[width=1\linewidth]{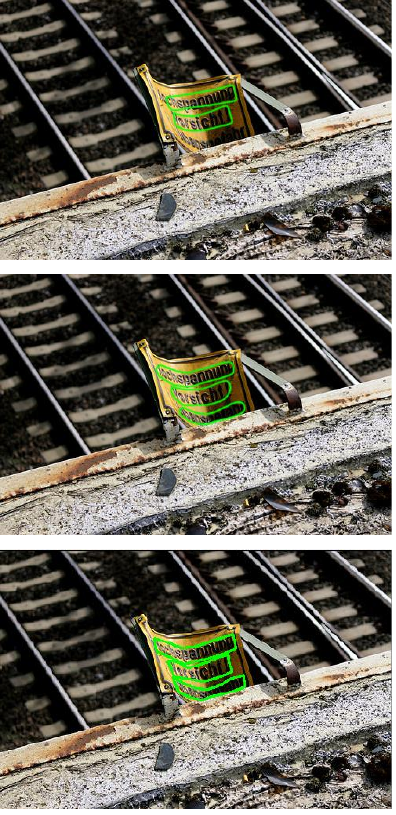}
	\end{minipage}}
	
	\caption{Qualitative comparisons with FCENet~\cite{zhu2021fourier}, TextField~\cite{xu2019textfield}, OPMP~\cite{zhang2020opmp}, and ABCNet~\cite{liu2020abcnet} on selected challenging samples in CTW1500. The first row shows the detection results of FCENet, TextField, OPMP, and ABCNet. The second row shows the detection results of ours. The last row shows the corresponding ground-truth.}
	\vspace{-4mm}
	\label{V11}
\end{figure*}

Moreover, we compared the computational cost of CM-Net with existing real-time detectors. To ensure a fair comparison, under the same setting, we employ ResNet18 as the backbone and resize the size of the input image as 640$\times$640$\times$3. As we can see from Table~\ref{speed2}, our method can reduce 44\% computational cost compared with PAN. Although DB uses lightweight FPN to reduce the computational cost of frameworks, CM-Net achieves better balance between detection accuracy and speed (as shown in Section~\ref{cwem}), which verifies the our method achieves the best comprehensive performance among existing real-time arbitrary-shaped text detection methods.

\textbf{Result Visualization.} We particularly show some qualitative results of highly-curved text instances in CTW1500 to demonstrate the superiority of our method for detecting hard irregular-shaped examples. As shown in Fig.~\ref{V11} and Table~\ref{ctw1500tt}, although FCENet~\cite{zhu2021fourier}, TextField~\cite{xu2019textfield}, OPMP~\cite{zhang2020opmp}, and ABCNet~\cite{liu2020abcnet} design complicated networks and post-processing to more effective detect irregular-shaped text instances, the detection results are still worse than our method. Specifically, FCENet fails to detect small-scale text instances (Fig.~\ref{V11}~(a)~first row). TextField performs badly for great aspect ratio text instances (Fig.~\ref{V11}~(b)~first row). OPMP and ABCNet can not discriminate the text instances from the background that the texture feature is similar to text instances (Fig.~\ref{V11}~(c),(d))~first row). These detection results demonstrate the effectiveness of the proposed CM-Net for irregular-shaped text detection.

\section{Conclusion}
In this work, we propose a novel efficient framework for detecting arbitrary-shaped text instances in high detection accuracy and speed simultaneously. The concentric mask is proposed to fit text contours in a more robust way compared with the shrink mask, which is helpful for our method to avoids the influences brought by text instance shapes. Moreover, we propose a multi-perspective feature (MPF) module to encourage CM-Net to learn more discriminative concentric mask-related features, which significantly improves the model ability to recognize concentric masks and brings no extra computational cost to the inference process. Additionally, a multi-factor constraints loss is proposed to supervise the training of CM-Net. With the aforementioned advantages, the proposed CM-Net achieves excellent performance in both detection speed and accuracy on multiple public text detection datasets. Experiments demonstrate that the proposed method outperforms all existing SOTA real-time methods by a large margin on four classic public benchmarks.

\ifCLASSOPTIONcaptionsoff
  \newpage
\fi



\bibliographystyle{IEEEtran}
\bibliography{egbib}

\begin{thebibliography}{10}
\providecommand{\url}[1]{#1}
\csname url@samestyle\endcsname
\providecommand{\newblock}{\relax}
\providecommand{\bibinfo}[2]{#2}
\providecommand{\BIBentrySTDinterwordspacing}{\spaceskip=0pt\relax}
\providecommand{\BIBentryALTinterwordstretchfactor}{4}
\providecommand{\BIBentryALTinterwordspacing}{\spaceskip=\fontdimen2\font plus
\BIBentryALTinterwordstretchfactor\fontdimen3\font minus
  \fontdimen4\font\relax}
\providecommand{\BIBforeignlanguage}[2]{{%
\expandafter\ifx\csname l@#1\endcsname\relax
\typeout{** WARNING: IEEEtran.bst: No hyphenation pattern has been}%
\typeout{** loaded for the language `#1'. Using the pattern for}%
\typeout{** the default language instead.}%
\else
\language=\csname l@#1\endcsname
\fi
#2}}
\providecommand{\BIBdecl}{\relax}
\BIBdecl

\bibitem{lin2017feature}
T.-Y. Lin, P.~Doll{\'a}r, R.~Girshick, K.~He, B.~Hariharan, and S.~Belongie,
  ``Feature pyramid networks for object detection,'' in \emph{Proceedings of
  the IEEE Conference on Computer Vision and Pattern Recognition}, 2017, pp.
  2117--2125.

\bibitem{lin2017focal}
T.-Y. Lin, P.~Goyal, R.~Girshick, K.~He, and P.~Doll{\'a}r, ``Focal loss for
  dense object detection,'' in \emph{Proceedings of the IEEE International
  Conference on Computer Vision}, 2017, pp. 2980--2988.

\bibitem{yuan2019vssa}
Y.~Yuan, Z.~Xiong, and Q.~Wang, ``Vssa-net: vertical spatial sequence attention
  network for traffic sign detection,'' \emph{IEEE transactions on image
  processing}, vol.~28, no.~7, pp. 3423--3434, 2019.

\bibitem{milletari2016v}
F.~Milletari, N.~Navab, and S.-A. Ahmadi, ``V-net: Fully convolutional neural
  networks for volumetric medical image segmentation,'' in \emph{2016 Fourth
  International Conference on 3D Vision}, pp. 565--571.

\bibitem{long2015fully}
J.~Long, E.~Shelhamer, and T.~Darrell, ``Fully convolutional networks for
  semantic segmentation,'' in \emph{Proceedings of the IEEE Conference on
  Computer Vision and Pattern Recognition}, 2015, pp. 3431--3440.

\bibitem{xiong2021ask}
Z.~Xiong, Y.~Yuan, and Q.~Wang, ``Ask: Adaptively selecting key local features
  for rgb-d scene recognition,'' \emph{IEEE Transactions on Image Processing},
  vol.~30, pp. 2722--2733, 2021.

\bibitem{he2017mask}
K.~He, G.~Gkioxari, P.~Doll{\'a}r, and R.~Girshick, ``Mask r-cnn,'' in
  \emph{Proceedings of the IEEE International Conference on Computer Vision},
  2017, pp. 2961--2969.

\bibitem{xie2020polarmask}
E.~Xie, P.~Sun, X.~Song, W.~Wang, X.~Liu, D.~Liang, C.~Shen, and P.~Luo,
  ``Polarmask: Single shot instance segmentation with polar representation,''
  in \emph{Proceedings of the IEEE Conference on Computer Vision and Pattern
  Recognition}, 2020, pp. 12\,193--12\,202.

\bibitem{zhang2019mask}
H.~Zhang, Y.~Tian, K.~Wang, W.~Zhang, and F.-Y. Wang, ``Mask ssd: an effective
  single-stage approach to object instance segmentation,'' \emph{IEEE
  Transactions on Image Processing}, vol.~29, pp. 2078--2093, 2019.

\bibitem{zhou2017east}
X.~Zhou, C.~Yao, H.~Wen, Y.~Wang, S.~Zhou, W.~He, and J.~Liang, ``East: an
  efficient and accurate scene text detector,'' in \emph{Proceedings of the
  IEEE Conference on Computer Vision and Pattern Recognition}, 2017, pp.
  5551--5560.

\bibitem{ma2018arbitrary}
J.~Ma, W.~Shao, H.~Ye, L.~Wang, H.~Wang, Y.~Zheng, and X.~Xue,
  ``Arbitrary-oriented scene text detection via rotation proposals,''
  \emph{IEEE Transactions on Multimedia}, vol.~20, no.~11, pp. 3111--3122,
  2018.

\bibitem{liao2018rotation}
M.~Liao, Z.~Zhu, B.~Shi, G.~Xia, and X.~Bai, ``Rotation-sensitive regression
  for oriented scene text detection,'' in \emph{Proceedings of the IEEE
  Conference on Computer Vision and Pattern Recognition}, 2018, pp. 5909--5918.

\bibitem{feng2019textdragon}
W.~Feng, W.~He, F.~Yin, X.-Y. Zhang, and C.-L. Liu, ``Textdragon: An end-to-end
  framework for arbitrary shaped text spotting,'' in \emph{Proceedings of the
  IEEE International Conference on Computer Vision}, 2019, pp. 9076--9085.

\bibitem{wang2020textray}
F.~Wang, Y.~Chen, F.~Wu, and X.~Li, ``Textray: Contour-based geometric modeling
  for arbitrary-shaped scene text detection,'' in \emph{Proceedings of the 28th
  ACM International Conference on Multimedia}, 2020, pp. 111--119.

\bibitem{liu2020abcnet}
Y.~Liu, H.~Chen, C.~Shen, T.~He, L.~Jin, and L.~Wang, ``Abcnet: Real-time scene
  text spotting with adaptive bezier-curve network,'' in \emph{Proceedings of
  the IEEE Conference on Computer Vision and Pattern Recognition}, 2020, pp.
  9809--9818.

\bibitem{wang2019efficient}
W.~Wang, E.~Xie, X.~Song, Y.~Zang, W.~Wang, T.~Lu, G.~Yu, and C.~Shen,
  ``Efficient and accurate arbitrary-shaped text detection with pixel
  aggregation network,'' in \emph{Proceedings of the IEEE International
  Conference on Computer Vision}, 2019, pp. 8440--8449.

\bibitem{liao2020real}
M.~Liao, Z.~Wan, C.~Yao, K.~Chen, and X.~Bai, ``Real-time scene text detection
  with differentiable binarization.'' in \emph{Proceedings of the AAAI
  Conference on Artificial Intelligence}, 2020, pp. 11\,474--11\,481.

\bibitem{zhang2020deep}
S.~Zhang, X.~Zhu, J.~Hou, C.~Liu, C.~Yang, H.~Wang, and X.~Yin, ``Deep
  relational reasoning graph network for arbitrary shape text detection,'' in
  \emph{Proceedings of the IEEE Conference on Computer Vision and Pattern
  Recognition}, 2020, pp. 9699--9708.

\bibitem{ma2021relatext}
C.~Ma, L.~Sun, Z.~Zhong, and Q.~Huo, ``Relatext: exploiting visual
  relationships for arbitrary-shaped scene text detection with graph
  convolutional networks,'' \emph{Pattern Recognition}, vol. 111, p. 107684,
  2021.

\bibitem{wang2020contournet}
Y.~Wang, H.~Xie, Z.~Zha, M.~Xing, Z.~Fu, and Y.~Zhang, ``Contournet: Taking a
  further step toward accurate arbitrary-shaped scene text detection,'' in
  \emph{Proceedings of the IEEE Conference on Computer Vision and Pattern
  Recognition}, 2020, pp. 11\,753--11\,762.

\bibitem{he2017single}
P.~He, W.~Huang, T.~He, Q.~Zhu, Y.~Qiao, and X.~Li, ``Single shot text detector
  with regional attention,'' in \emph{Proceedings of the IEEE International
  Conference on Computer Vision}, 2017, pp. 3047--3055.

\bibitem{liao2018textboxes++}
M.~Liao, B.~Shi, and X.~Bai, ``Textboxes++: A single-shot oriented scene text
  detector,'' \emph{IEEE Transactions on Image Processing}, vol.~27, no.~8, pp.
  3676--3690, 2018.

\bibitem{liu2019omnidirectional}
Y.~Liu, S.~Zhang, L.~Jin, L.~Xie, Y.~Wu, and Z.~Wang, ``Omnidirectional scene
  text detection with sequential-free box discretization,'' \emph{arXiv
  preprint arXiv:1906.02371}, 2019.

\bibitem{tang2019seglink++}
J.~Tang, Z.~Yang, Y.~Wang, Q.~Zheng, Y.~Xu, and X.~Bai, ``Seglink++: Detecting
  dense and arbitrary-shaped scene text by instance-aware component grouping,''
  \emph{Pattern Recognition}, vol.~96, p. 106954, 2019.

\bibitem{liu2016ssd}
W.~Liu, D.~Anguelov, D.~Erhan, C.~Szegedy, S.~Reed, C.~Fu, and A.~Berg, ``Ssd:
  Single shot multibox detector,'' in \emph{Proceedings of the European
  Conference on Computer Vision}, 2016, pp. 21--37.

\bibitem{zhang2019look}
C.~Zhang, B.~Liang, Z.~Huang, M.~En, J.~Han, E.~Ding, and X.~Ding, ``Look more
  than once: An accurate detector for text of arbitrary shapes,'' in
  \emph{Proceedings of the IEEE Conference on Computer Vision and Pattern
  Recognition}, 2019, pp. 10\,552--10\,561.

\bibitem{wang2020all}
H.~Wang, P.~Lu, H.~Zhang, M.~Yang, X.~Bai, Y.~Xu, M.~He, Y.~Wang, and W.~Liu,
  ``All you need is boundary: Toward arbitrary-shaped text spotting,'' in
  \emph{Proceedings of the AAAI Conference on Artificial Intelligence},
  vol.~34, no.~07, 2020, pp. 12\,160--12\,167.

\bibitem{zhu2021fourier}
Y.~Zhu, J.~Chen, L.~Liang, Z.~Kuang, L.~Jin, and W.~Zhang, ``Fourier contour
  embedding for arbitrary-shaped text detection,'' \emph{arXiv preprint
  arXiv:2104.10442}, 2021.

\bibitem{lyu2018mask}
P.~Lyu, M.~Liao, C.~Yao, W.~Wu, and X.~Bai, ``Mask textspotter: An end-to-end
  trainable neural network for spotting text with arbitrary shapes,'' in
  \emph{Proceedings of the European Conference on Computer Vision}, 2018, pp.
  67--83.

\bibitem{long2018textsnake}
S.~Long, J.~Ruan, W.~Zhang, X.~He, W.~Wu, and C.~Yao, ``Textsnake: A flexible
  representation for detecting text of arbitrary shapes,'' in \emph{Proceedings
  of the European Conference on Computer Vision}, 2018, pp. 20--36.

\bibitem{baek2019character}
Y.~Baek, B.~Lee, D.~Han, S.~Yun, and H.~Lee, ``Character region awareness for
  text detection,'' in \emph{Proceedings of the IEEE Conference on Computer
  Vision and Pattern Recognition}, 2019, pp. 9365--9374.

\bibitem{dai2016r}
J.~Dai, Y.~Li, K.~He, and J.~Sun, ``R-fcn: Object detection via region-based
  fully convolutional networks,'' \emph{arXiv preprint arXiv:1605.06409}, 2016.

\bibitem{lyu2018multi}
P.~Lyu, C.~Yao, W.~Wu, S.~Yan, and X.~Bai, ``Multi-oriented scene text
  detection via corner localization and region segmentation,'' in
  \emph{Proceedings of the IEEE Conference on Computer Vision and Pattern
  Recognition}, 2018, pp. 7553--7563.

\bibitem{zhang2020opmp}
S.~Zhang, Y.~Liu, L.~Jin, Z.~Wei, and C.~Shen, ``Opmp: An omnidirectional
  pyramid mask proposal network for arbitrary-shape scene text detection,''
  \emph{IEEE Transactions on Multimedia}, vol.~23, pp. 454--467, 2020.

\bibitem{wang2019shape}
W.~Wang, E.~Xie, X.~Li, W.~Hou, T.~Lu, G.~Yu, and S.~Shao, ``Shape robust text
  detection with progressive scale expansion network,'' in \emph{Proceedings of
  the IEEE Conference on Computer Vision and Pattern Recognition}, 2019, pp.
  9336--9345.

\bibitem{tian2019learning}
Z.~Tian, M.~Shu, P.~Lyu, R.~Li, C.~Zhou, X.~Shen, and J.~Jia, ``Learning
  shape-aware embedding for scene text detection,'' in \emph{Proceedings of the
  IEEE Conference on Computer Vision and Pattern Recognition}, 2019, pp.
  4234--4243.

\bibitem{jiang2020arbitrary}
X.~Jiang, S.~Xu, S.~Zhang, and S.~Cao, ``Arbitrary-shaped text detection with
  adaptive text region representation,'' \emph{IEEE Access}, vol.~8, pp.
  102\,106--102\,118, 2020.

\bibitem{xu2019textfield}
Y.~Xu, Y.~Wang, W.~Zhou, Y.~Wang, Z.~Yang, and X.~Bai, ``Textfield: Learning a
  deep direction field for irregular scene text detection,'' \emph{IEEE
  Transactions on Image Processing}, vol.~28, no.~11, pp. 5566--5579, 2019.

\bibitem{he2016deep}
K.~He, X.~Zhang, S.~Ren, and J.~Sun, ``Deep residual learning for image
  recognition,'' in \emph{Proceedings of the IEEE Conference on Computer Vision
  and Pattern Recognition}, 2016, pp. 770--778.

\bibitem{yao2012detecting}
C.~Yao, X.~Bai, W.~Liu, Y.~Ma, and Z.~Tu, ``Detecting texts of arbitrary
  orientations in natural images,'' in \emph{IEEE Conference on Computer Vision
  and Pattern Recognition}, 2012, pp. 1083--1090.

\bibitem{yao2014unified}
C.~Yao, X.~Bai, and W.~Liu, ``A unified framework for multioriented text
  detection and recognition,'' \emph{IEEE Transactions on Image Processing},
  vol.~23, no.~11, pp. 4737--4749, 2014.

\bibitem{yuliang2017detecting}
Y.~Liu, L.~Jin, S.~Zhang, and S.~Zhang, ``Detecting curve text in the wild: New
  dataset and new solution,'' \emph{arXiv preprint arXiv:1712.02170}, 2017.

\bibitem{ch2017total}
C.~K. Ch'ng and C.~S. Chan, ``Total-text: A comprehensive dataset for scene
  text detection and recognition,'' in \emph{Proceedings of the International
  Conference on Document Analysis and Recognition}, vol.~1, 2017, pp. 935--942.

\bibitem{karatzas2015icdar}
D.~Karatzas, L.~Gomez, A.~Nicolaou, S.~Ghosh, A.~Bagdanov, M.~Iwamura,
  J.~Matas, L.~Neumann, V.~Chandrasekhar, and S.~Lu, ``Icdar 2015 competition
  on robust reading,'' in \emph{Proceedings of the International Conference on
  Document Analysis and Recognition}, 2015, pp. 1156--1160.

\bibitem{kingma2014adam}
D.~Kingma and J.~Ba, ``Adam: A method for stochastic optimization,''
  \emph{arXiv preprint arXiv:1412.6980}, 2014.

\bibitem{yu2018bisenet}
C.~Yu, J.~Wang, C.~Peng, C.~Gao, G.~Yu, and N.~Sang, ``Bisenet: Bilateral
  segmentation network for real-time semantic segmentation,'' in
  \emph{Proceedings of the European Conference on Computer Vision}, 2018, pp.
  325--341.

\bibitem{dai2017deformable}
J.~Dai, H.~Qi, Y.~Xiong, Y.~Li, G.~Zhang, H.~Hu, and Y.~Wei, ``Deformable
  convolutional networks,'' in \emph{Proceedings of the IEEE International
  Conference on Computer Vision}, 2017, pp. 764--773.

\bibitem{gupta2016synthetic}
A.~Gupta, A.~Vedaldi, and A.~Zisserman, ``Synthetic data for text localisation
  in natural images,'' in \emph{Proceedings of the IEEE Conference on Computer
  Vision and Pattern Recognition}, 2016, pp. 2315--2324.

\bibitem{deng2018pixellink}
D.~Deng, H.~Liu, X.~Li, and D.~Cai, ``Pixellink: Detecting scene text via
  instance segmentation,'' in \emph{Proceedings of the AAAI Conference on
  Artificial Intelligence}, vol.~32, no.~1, 2018.

\bibitem{rosenfeld1971edge}
A.~Rosenfeld and M.~Thurston, ``Edge and curve detection for visual scene
  analysis,'' \emph{IEEE Transactions on Computers}, vol. 100, no.~5, pp.
  562--569, 1971.

\end{thebibliography}
%
%





\end{document}